\documentclass[sigconf,authorversion,nonacm]{acmart}

\acmConference[UIST 'XX]{ACM Symposium on User Interface Software and Technology}{Month XX–XX, 20XX}{Location}
\acmBooktitle{Proceedings of the ACM Symposium on User Interface Software and Technology (UIST 'XX), Month XX–XX, 20XX, Location}
\acmDOI{XX.XXXX/XXXXXXX.XXXXXXX}
\acmISBN{978-1-XXXX-XXXX-X/XX/XX}

\usepackage{graphicx}
\usepackage{subcaption}
\usepackage{booktabs}
\usepackage{microtype}
\usepackage{balance}
\usepackage{hyperref}
\usepackage{natbib}
\usepackage{algorithm}
\usepackage{algorithmic}
\usepackage{enumitem}
\usepackage{tabularx}
\usepackage{array}  
\usepackage{amsthm}
\usepackage{lipsum}
\usepackage{arydshln}
\usepackage{amsmath}
\usepackage{pdflscape}
\usepackage{longtable}
\usepackage{placeins}

\newcommand{\model}{\textsc{POET}}
\newcommand{\modelone}{\textsc{Base}}
\newcommand{\modeltwo}{\textsc{POET}}
\newcommand{\modelthree}{\textsc{Base-Personalize}}
\newcommand{\modelfour}{\textsc{POET-Personalize}}

\DeclareMathOperator*{\argmax}{arg\,max}

\newcommand{\revised}[1]{\textcolor{black}{#1}}

\title{Anonymous Submission}

\author{Evans Xu Han}
\affiliation{%
  \institution{Stanford University}
  \city{Palo Alto}
  \state{CA}
  \country{USA}
}
\affiliation{%
  \institution{Yale University}
  \city{New Haven}
  \state{CT}
  \country{USA}
}
\email{evanshan@cs.stanford.edu}

\author{Alice Qian Zhang}
\affiliation{%
  \institution{Carnegie Mellon University}
  \city{Pittsburgh}
  \state{PA}
  \country{USA}
}
\email{aqzhang@andrew.cmu.edu}

\author{Haiyi Zhu}
\affiliation{%
  \institution{Carnegie Mellon University}
  \city{Pittsburgh}
  \state{PA}
  \country{USA}
}
\email{haiyiz@andrew.cmu.edu}

\author{Hong Shen}
\affiliation{%
  \institution{Carnegie Mellon University}
  \city{Pittsburgh}
  \state{PA}
  \country{USA}
}
\email{hongs@andrew.cmu.edu}

\author{Paul Pu Liang}
\affiliation{%
  \institution{Massachusetts Institute of Technology}
  \city{Cambridge}
  \state{MA}
  \country{USA}
}
\email{ppliang@mit.edu}

\author{Jane Hsieh}
\affiliation{%
  \institution{Carnegie Mellon University}
  \city{Pittsburgh}
  \state{PA}
  \country{USA}
}
\email{jhsieh2@andrew.cmu.edu}

\def\plaintitle{\model: 
Supporting Prompting Creativity and Personalization with Automated Expansion of Text-to-Image Generation
}

\begin{document}

\title[POET]{\plaintitle}

\begin{abstract}

% \todo{can we make sure all figures and tables are referred to in the text}

State-of-the-art visual generative AI tools hold immense potential to assist users in the early ideation stages of creative tasks --- offering the ability to generate (rather than search for) novel and unprecedented (instead of existing) images of considerable quality that also adhere to boundless combinations of user specifications.
However, many large-scale text-to-image systems are designed for broad applicability, yielding conventional output that may limit creative exploration. They also employ interaction methods that may be difficult for beginners.
Given that creative end-users often operate in diverse, context-specific ways that are often unpredictable, more variation and personalization are necessary.
We introduce \model, a real-time interactive tool that (1) automatically discovers dimensions of homogeneity in text-to-image generative models, (2) expands these dimensions to diversify the output space of generated images, and (3) learns from user feedback to personalize expansions.
An evaluation with 28 users spanning four creative task domains demonstrated \model's ability to generate results with higher perceived diversity and help users reach satisfaction in fewer prompts during creative tasks, thereby prompting them to deliberate and reflect more on a wider range of possible produced results during the co-creative process.
Focusing on visual creativity, \model\ offers a first glimpse of how interaction techniques of future text-to-image generation tools may support and align with more pluralistic values and the needs of end-users during the ideation stages of their work\footnote{The project page with code can be found at \url{https://github.com/evansh666/POET}.}.
\end{abstract}

\begin{CCSXML}
<ccs2012>
   <concept>
       <concept_id>10003120.10003121.10003129</concept_id>
       <concept_desc>Human-centered computing~Interactive systems and tools</concept_desc>
       <concept_significance>500</concept_significance>
       </concept>
   <concept>
       <concept_id>10010147.10010178</concept_id>
       <concept_desc>Computing methodologies~Artificial intelligence</concept_desc>
       <concept_significance>500</concept_significance>
       </concept>
 </ccs2012>
\end{CCSXML}

\ccsdesc[500]{Human-centered computing~Interactive systems and tools}
\ccsdesc[500]{Computing methodologies~Artificial intelligence}

\keywords{Text-to-image generation, creativity, personalization, pluralism} %would plurality be better?

% moving to front of paper for impact -- can move around & iterate on for final looks!
\begin{teaserfigure}
    \centering
    \includegraphics[width=1\textwidth]{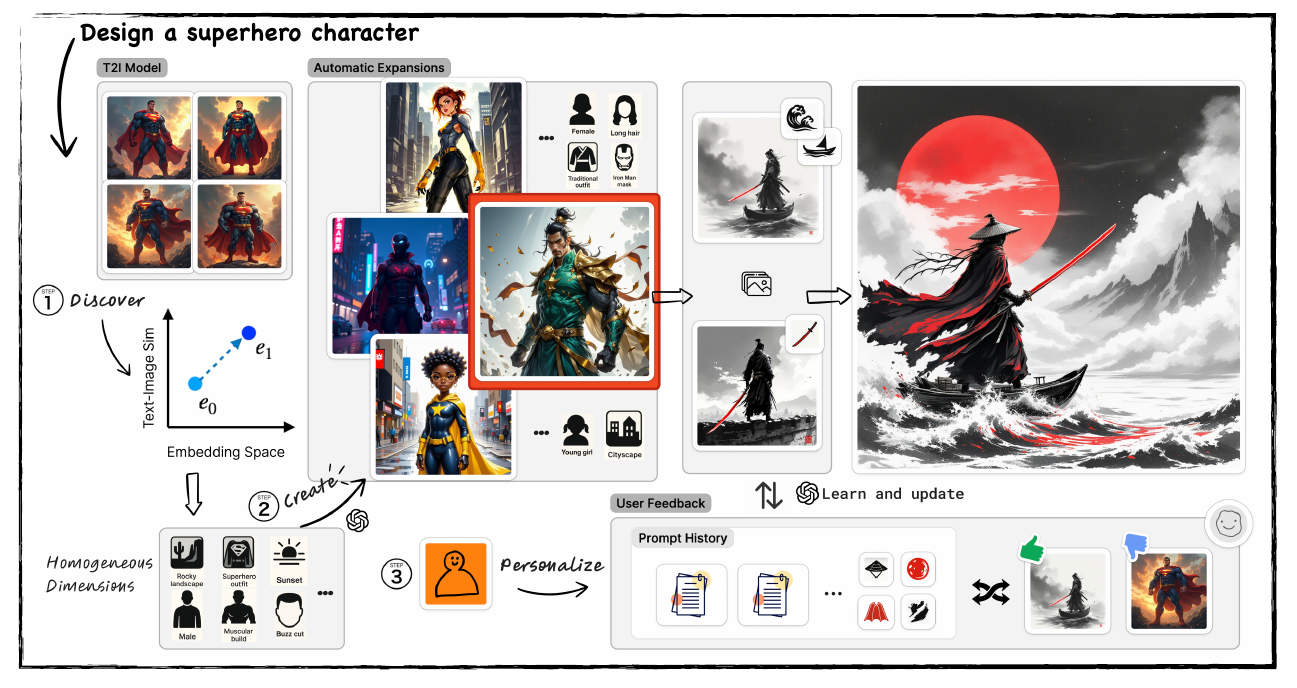}
    \caption{Despite advances in generative AI, current output spaces and prompting workflows are not designed to accommodate the diverse, evolving, and unpredictable practices, contexts, and preferences of visual artists and designers. We introduce \model: a text-to-image generation tool that supports creative tasks by diversifying outputs along automatically discovered dimensions, while maintaining fidelity and consistency. \model\ (1) automatically discovers dimensions of homogeneity in text-to-image generative models, (2) expands these dimensions to diversify the output space of generated images, and (3) learns from user feedback to personalize expansions.}
    \label{fig:enter-label}
\end{teaserfigure}

\maketitle
\section{Introduction}
Generative AI tools are revolutionizing creative domains such as design \cite{3dalle, architectural, interior}, visual art \cite{art}, marketing \cite{genmarketing, productstyling, product_education}, game character art \cite{sketchar}, fashion \cite{fashion, fashion_midjourney}, and journalism \cite{liu2022opal}, driving a paradigm shift for creative production processes. 
In particular, the advent of large-scale text-to-image generation models such as Stable Diffusion \cite{sd} and DALL-E \cite{dalle} provides impressive capabilities to produce high-quality results, opening avenues for creative professionals to augment their existing workflows to ideate and produce with assistance from generative AI systems \cite{ltgms}. Crucially, these models can generate images in a zero-shot fashion, using only text and/or image input \cite{zeroshot,liang2024foundations}, enabling users from various creative domains to visualize and ideate concepts with ease. For creative tasks that cater to wide-ranging audiences, text-to-image generation tools are shown to augment human creativity and produce better-received content \cite{genmarketing, productivity}.

However, despite their powerful generative capabilities,  current output spaces and prompting workflows of generative AI systems are not designed to accommodate the diverse \cite{diverse}, evolving \cite{evolving}, and unpredictable \cite{unpredictable} practices \cite{plurality}, contexts \cite{contextcam, demo} and preferences \cite{plural_alignment} of visual artists and designers. Despite their prevalence and wide range of applicable domains, novice users struggle to integrate generative systems into their creative workflows \cite{news_challenge} or view them as constraints to creativity \cite{hindrance}. Critical scholars also voice concerns about how these models reflect and amplify the normative power of their founding institutions over end-users \cite{unstraightening, positivity} -- influencing not only content, but the %achieved by structuring users' 
creative processes \cite{power} and aesthetics preferences of 
users \cite{aiaesthetics}. \revised{Prior works with dancers \cite{dancer} also find creative practitioners often struggle to use generative models due to their conventional and limited output space. They recognize the limitations of these models and express a desire for new ones that can produce \textit{less clichéd} art to better augment their creative processes~\citep{glitch, uncertainty, positivity}.}
% , and possibly disregarding copyright of creators to challenge those criticizing their approach \cite{vulgar}. 

Across HCI, Machine Learning, and Computer Vision communities, researchers increasingly advocate for interaction and evaluation techniques that serve more pluralistic values of end-user communities that leverage generative tools \cite{plural_alignment, plurality, hifi, urban}. In response, recent work approaching this gap explored ways of improving alignment with user expectations: prompt refinement through multi-modal feedback \cite{charm} or reinforcement learning \cite{rl_optimizing}, synthetic prompt mapping to visualize the search space \cite{promptmap} and interactions to achieve personalized style alignment \cite{stylefactory}. 
However, such techniques focused mainly on changing and optimizing user prompts to broaden the design space -- overlooking opportunities to (1) defy the normative values and stereotypes that generative models are found to reflect and amplify from training data \cite{amplify, gstereotypes, unstraightening} and (2) alleviate user burdens of re-prompting to clarify intentions.

%\todo{should we say stereotypes? and can we expand on point 2 to make stronger, seems a critical transition to motivate our work}
% But the focus of such works on aligning to user needs and intentions circumvents opportunities to support the parts of creative practices that are intentionally effortful and endeavoring --- especially ones that resist and deliberate on established normative aesthetics and ideals \cite{power}. 
% While these 
% -- automatically identified and conducts targeted expansion output space of image generation, as opposed to making users do this through prompt expansion, to retain fidelity to original intentions of prompt, (balancing the trade-offs between expansion and fidelity/consistency)
% --- balance of prompt fidelity with output expansion
% % --- expansion will enable better personalization

To expand the output spaces of text-to-image models and to reduce the burden on users to manually refine prompts, we investigate ways to \textit{automatically detect and expand their homogeneous dimensions}, so that users can explore more varied design alternatives with less effort. However, while targeted expansions of output spaces are faster and less demanding in terms of user effort, such automatic methods are inherently at tension with prompt fidelity and, more generally, user agency.
To empower creative users with the convenience of automatically expanded output spaces while accounting for their expressed intent and preferences, we developed \model: a system for \underline{P}ersonalized \underline{O}utput \underline{E}xpansion for \underline{T}ext-to-image generation, which introduces a method for prompt inversion that (1) automatically discovers novel and potentially homogeneous dimensions (2) selectively changes the prompt into expanded versions along such dimensions and uses updated prompts to generate more desirable images and (3) learns from user feedback to personalize these expansions.
%\model~ leverages a recent text-to-image system with improved compositionality -- a criteria imperative for achieving fidelity to prompts ~\citep{han2024progressive} -- to maintain the prompt specifications while maximizing the diversity of generated images with under-specified attributes, thereby preserving user-specified input requirements while automatically expanding and learning more novel and pluralistic sets of user preferences. \todo{not sure we need this last sentence.}
% Through \model, we propose 
% This system will be trained to maintain the same information in the prompt while maximizing the diversity of generated images with under-specified attributes. 
% Maximizing this objective function by computing the gradient with changes in continuous prompt space will yield a model that automatically expands and learns from diverse user preferences. 
% For the underlying text-to-image model, we plan to experiment with a recent system we developed that achieves the best compositionality compared to other baselines, which is critical to generate images that carefully follow the prompt. 
\vspace{1em}

In summary, this study makes the following contributions:
\begin{itemize}
    \item \model, a text-to-image generation tool, \revised{operating at the ideation stage}, that supports creative tasks by diversifying and expanding outputs along automatically discovered dimensions while maintaining fidelity and consistency with user intent.
    \item A multi-step user study to evaluate 
    \begin{enumerate}
        \item the diversity of generated image outputs across five creative task domains
        \item its ability to support the creative prompting process
        \item its effectiveness in personalizing to users' preferences
    \end{enumerate}
\end{itemize}

\section{Related Work}
\label{related}

\begin{figure*}[t]
\vspace{1em} 
    \centering 
    \includegraphics[width=1\textwidth]{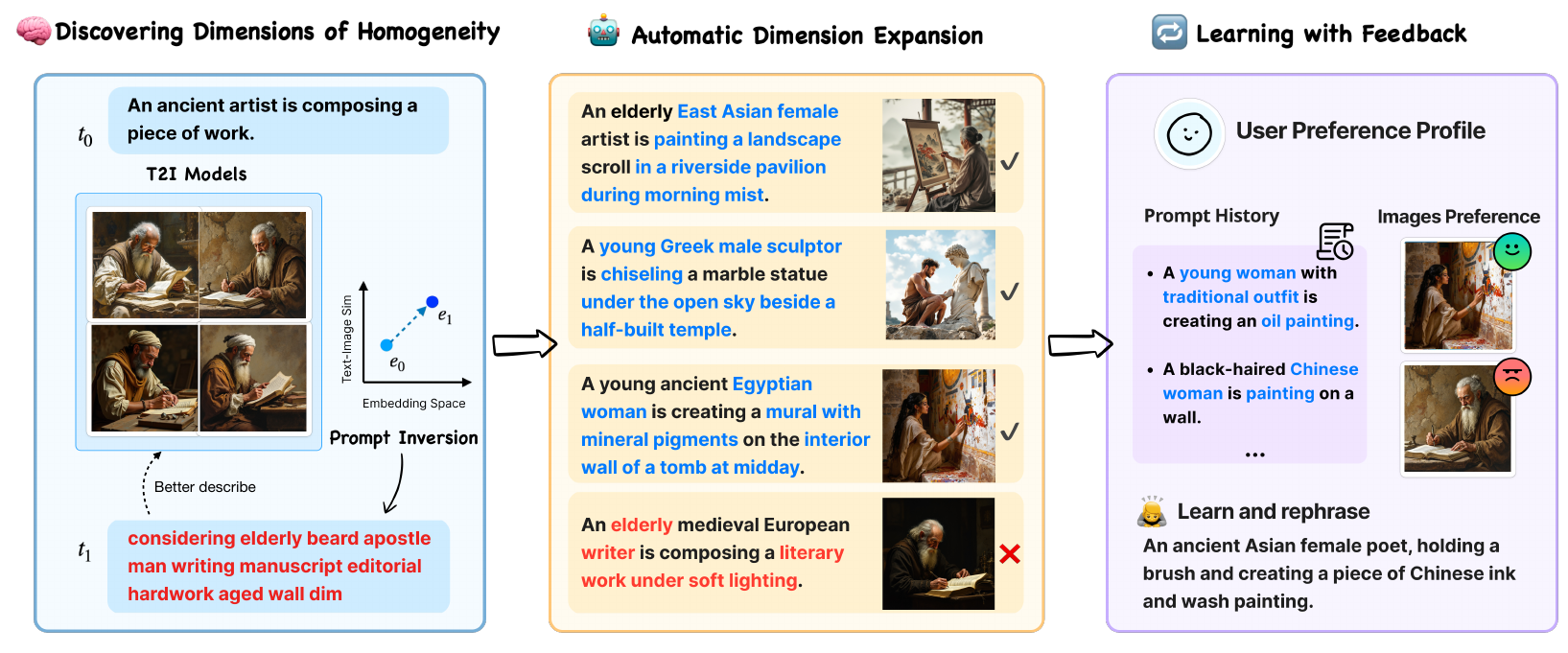}
    \caption{Image generation pipeline of our proposed system \model, which starts by discovering dimensions of homogeneity $t_1$ from a set of generated images from pretrained T2I models. This discovery is performed by a new prompt inversion method we developed that learns new token embeddings that are maximally similar to features present in all images.
    \model\ then expands the prompt along prioritized dimensions ($t_1$) to generate more diverse options before refining future expansions based on user feedback.}
    \label{fig:pipeline}
\end{figure*}

\subsection{Expanding Image Generation for Ideation, Co-creation \& Pluralistic Empowerment}
% this subsection draws on the creativity support literature, highlighting the potentials of t2i models to support end-users in early ideation processes, while also identifying the gap of how normative results can disempower users and creators
Across creative domains, studies suggest early evidence that generative AI systems can assist and augment early ideation stages of designers' creative processes \cite{product_ideation, fashion_midjourney, aideation, augment, creativeconnect, supermind}, engaging them in co-creativity --- where the creativity of the AI system and its user blends in the production process \cite{cocreativity}. 
% In various visual industries --- e.g., interior design \cite{interior}, game character art \cite{sketchar}, fashion \cite{fashion, fashion_midjourney}, product marketing design \cite{product_education} and cinematic composition \cite{cinematic} --- 
However, the integration of pre-trained visual generative models into creative workflows poses several challenges, arising from their unpredictability \cite{unpredictable} (and thus uncontrollability \cite{uncontrollable}), as well as the inheritance of human biases from training data \cite{doom}. 
Despite (or perhaps due to) the unpredictable nature of visual generative models, many creative professionals engage with them in a co-creative process to (1) harness their unique ability to generate novel images that have never been seen before \cite{inversion} so as to gather ideas and inspiration in early stages of creation \cite{architectural, fashion, product_ideation, aideation, supermind}, as well as to (2) experiment with their unexpected glitches and idiosyncrasies \cite{glitch, uncertainty}. \citet{power} characterizes creative practitioners' intentional use cases of visual generative tools as empowering, since it enables them to ``capture and interpret their perception of the world and share their unique perspectives with others to affect change''. For creative applications, the ability to generate unique and novel designs and ideas is even more imperative than traditional contexts of use.
However, because modern text-to-image generation models (e.g., Stable Diffusion \cite{sd} and DALL-E \cite{dalle}) are trained on large corpora of image-caption pairs from the Internet to accommodate as many domains and contexts as possible ~\citep{liang2024foundations}, their applications are broad and ill-suited to meet the task demands of more niche and non-standard tasks found in creative professions \cite{unstraightening, personalizing}. 

To help creative professionals fully harness the potential of models' unpredictability for co-creation, prior works explored ways of balancing its inherent tradeoffs to uncontrollability --- by surfacing and visualizing the prompt space for users \cite{promptmap}, refining textual prompts with multi-modal feedback \cite{charm}, or using GPT to generate (1) code that ``sketches out'' graphical inputs to guide the text-to-image generation \cite{controllable} or (2) more semantically diverse prompts \cite{designaid}. But while these approaches afford creative professionals the agency to explore and expand their prompt spaces during their use of text-to-image tools, it remains a challenge to address the unseen and unsurfaced uniformities, stereotypes and homogeneities that often occur from the model itself \cite{gstereotypes, chinchure2024tibet, dalle_bias, unstraightening}. Responding to calls for more pluralistic alignment \cite{plural_alignment, plurality}, scholars from the HCI and AI communities explored and documented approaches to quantifying and mitigating social biases and homogeneities in large language models 
\cite{liang2021towards,nadeem2020stereoset,sheng2019woman}, image retrieval models~\cite{hendricks2018women,otterbacher2018addressing}, and image generation models~\citep{cho2023dall, dalle_bias,social_t2i,chinchure2024tibet, risks}, as well as algorithmic systems more broadly \cite{shen2021everyday,fan2024user}. But while such related work investigated possible techniques for automatically detecting biases in text-to-image models \cite{chinchure2024tibet,social_t2i,dalle_bias, social, openbias}, it remains an open question whether it's possible to automatically expand upon discovered dimensions, without sacrificing consistency and user fidelity. In this work, we take a stride beyond prompt variations to propose automatic identification \textit{and} expansion of homogeneous dimensions that are prevalent among text-to-image models, so as to enable production of more visually diverse outputs and to support the imagination, inspiration, and creative exploration of end-users. \revised{{\model} operates at the ideation stage of creative end-users’ workflows, where it helps produce more imaginative initial concepts automatically. In contrast, systems like PromptPaint \citep{luminate} influences final outputs, which users can leverage to subsequently refine ideas brainstormed with \model. Prompt suggestion tools \cite{promptify, luminate, designaid} support ideation by extending to related prompts, but often require multiple interaction rounds and fall short in automatically expanding the visual solution space. }

\subsection{Personalization \& Prompting}
While generating varied and diverse output is important for inspiration and creativity during ideation, such techniques have inherent trade-offs to image quality and controllability \cite{StyleGAN, structural_consistency}. As the use of text-to-image generation tools becomes a mainstream phenomenon in creative sectors and beyond, creative professionals like artists increasingly fear the loss of ownership and agency over their productions at work \cite{aiartcontroversy, perceptions}.
A substantial body of work sought to improve the personalization capabilities of these text-to-image models: enabling them to understand latent user intents~\citep{wilf2023think}, giving users fine-grained control over such latent attributes \cite{GAC, pixart, lyu2022dime}, and aligning with user-provided concepts from text prompts \cite{attndreambooth} or images \cite{inversion, contextcam}. However, such studies focused broadly on improving model capabilities without close examinations of how feedback-based user interactions can help various creative professionals in the early ideation stages of their work.

% The increasing adoption of generative AI models are disrupting the creative sectors, reshaping long-held practices, redefining legal determinations around copyright, and contributing to practitioner concerns of job displacement in both creative writing (e.g., screenwriters \cite{screen}, comedians \cite{comedy}) and visual industries --- e.g., interior design \cite{interior}, game character art \cite{sketchar}, fashion \cite{fashion, fashion_midjourney}, product marketing design \cite{product_education}, cinematic composition \cite{cinematic}. Across creative domains, studies suggest early evidence that generative AI systems can assist and augment early ideation stages of designers' creative processes \cite{product_ideation, fashion_midjourney, aideation}, engaging them in co-creativity --- where the creativity of the AI system and its user blends in the production process \cite{cocreativity}. Many professionals of creative domains engage with models in the co-creative process to harness the unique ability of large text-to-image generation models to generate novel images that have never been seen before, and even at times to experiment with their idiosyncrasies, glitches and uncertainties \cite{glitch, uncertainty}.

% copyright controversies considered exploitation of artists \cite{aiartcontroversy}
 
Prompt engineering is one form of interaction that creative professionals engage in to control the style and quality of generated outputs, and is a creative process of its own \cite{oppenlaenderart, t2i_creativity}. However, current prompting interactions for text-to-image generation are not designed to take into account the preferences and feedback of creative professionals \cite{demo}, even though techniques such as the use of prompt modifiers \cite{oppenlaender2024taxonomy} show the potential to mitigate traditional forms of biases embedded in AI systems. But while personalized prompting has been explored in the text-based space of NLP \cite{personalized, prompt_learning, LLaVA}, few (real-time interactive) systems explore the potential of catering to users' \textit{visual} preferences \cite{InstantBooth, imagine}, and even fewer aligns to task specific needs of creative end-users.

\revised{\subsection{Soft Prompt Optimization}
Soft prompt optimization aims to refine representations of user prompts with added words from the embedding space of a frozen text-to-image model. A prominent line of work in this direction is textual inversion \cite{inversion, inversion2}, which augments prompts with learned vectors to better steer image generation. While effective, these methods primarily operate on a per-prompt or per-image basis and are limited in their ability to generalize across multiple generations. Recent work~\cite{mo2024dynamicpromptoptimizingtexttoimage, mahajan2023promptinghardhardlyprompting} has optimized the per-image generation (denoising) process, but did not extend to multi-image optimization. 
% POET generalizes to group-level multi-image optimization, uniquely enabling automatic detection of homogeneous dimensions. 
Other approaches \cite{inversion2} improved overall task performance (e.g., style transfer, distillation), but these works did not explore how automatic detections of visual homogeneities can support real-time, user-driven creativity. Finally, prior studies optimized on discrete rather than continuous spaces~\cite{mo2024dynamicpromptoptimizingtexttoimage, mahajan2023promptinghardhardlyprompting}, overlooking the opportunity to offer further interpretability.}

% POET auto-diversifies to support user-driven creativity in real time, and  

\section{The \model\ System Architecture \& Capabilities} \label{system}

\paragraph{Target Creative User Communities} A creativity support tool -- as characterized by \citet{cst} -- ``runs on one or more digital systems, encompasses one or more creativity-focused features, and is employed to positively influence users of varying expertise in one or more distinct phases of the creative process.'' 
In this work, \model~ contributes to the early (e.g., ideation) stages of a user's creative task, where the target end-user population comprises professionals with expertise in a creative domain (i.e., product/destination marketer, game design, interior design). 
While these professionals include experts from varying creative domains, our tool is primarily designed for novice users of generative text-to-image systems since expert prompt engineers or AI artists may prefer to circumvent our systems' goals of expanded output generation and personalization or achieve them through their own developed methods.

Below, we describe the three main steps underlying \model's key features of output expansion and personalization, as well as the four text-to-image generation models that form the conditions of our experiment (detailed in \S\ref{exp_design}).
% Starting with  $symbol?$ and consisting of three steps. 
\begin{itemize}
    \item \modelone: In the first two phases of our experiment, the baseline model consists of the pre-trained text-to-image generative model Flux.1-dev (note that this backbone can be swapped with any open-weight T2I model).
    
    \item \modeltwo: Our contribution is \model, an optimized model that \textit{identifies} common dimensions of homogeneity across $n$ generated images from a base model and \textit{expands} these homogeneous dimensions to automatically add variation to the set of generated images.
    
    \item \modelthree: Our secondary contribution is an additional step of \textit{personalization}, where text-to-image models can learn from the user by taking as input images that they expressed preference (or aversion for) in a previous iteration, conditioning on a sequence of prompts previously explored, and iteratively improving its expansion strategy while aligning more closely with user's intent. We apply this personalization step to a base Flux.1-dev model for the Personalized-Baseline.  

    \item \modelfour: Finally, we apply this personalization step to our \model\ model to combine the power of homogeneous dimension identification and expansion with personalized learning from user feedback on expanded dimensions.
\end{itemize}

\subsection{Identifying Dimensions of Homogeneity}

Consider the case where a user inputs an initial prompt $\mathbf{t_0}$ to a standard text-to-image model (e.g., Stable Diffusion, Dall-E, Flux) to generate $n$ images -- the resulting output images would often share overly similar characteristics, reflecting homogeneous dimensions learned by the model. For example, the model might generate images with overwhelmingly homogeneous demographics, backgrounds, and locations. The first step of \model\ is to automatically \textit{identify} these common dimensions of homogeneity across the $n=10$ generated images so that they can be appropriately expanded and diversified. This can be a challenge since these dimensions are unknown a priori and unknown to pre-trained vision-language models (e.g., LLaVa~\citep{liu2023llava} or BLIP-2~\citep{blip2}), which makes direct prompting difficult. Furthermore, when the set of images is large, direct prompting for homogeneous dimensions can be computationally expensive to discover.
%because pre-trained image-captioning models or most vision-language models (e.g., LLaVa, BLIP-2) \textit{lack cross-image awareness} as these models are primarily optimized for single-image input and are not designed to reason over a set of images. Even when multi-image input is supported, these models often struggle to describe the commonalities within a group of 10 images simultaneously, and inference can be computationally expensive. Alternatively, more capable multi-image models (e.g., GPT-4v) can better handle joint visual reasoning over multiple images--but are closed-sourced and costly to access, making them impractical for large-scale automated analysis. Moreover, such systems often inherit biases from their training data, making them prone to overlooking or even reinforcing the biases we seek to identify. For example, such models may consistently conclude that the images exhibit specific social, demographic, or gender biases—regardless of whether these patterns are actually significant—due to their own training-induced priors. [\todo ablation studies about comparing those with ours. 1. inference time, 2. result quality]

\begin{algorithm}[t!]
   \caption{Prompt inversion algorithm to discover homogeneous dimensions.}
   \label{alg:po}
\begin{algorithmic}[]
\STATE \textbf{Input:} Model $\theta$, vocabulary embedding $\mathbf{E}^{|V|}$, original prompt $\mathbf{t_0} = [w_1, ..., \revised{w_n}]$, projection function $E^{-1}_{\mathbf{E}}$, objective function $\mathcal{L}$, optimization steps $T$, batch size $b$, learning rate $\gamma$, Image set $\mathbf{I}$.
\STATE {\color{gray} Initialize from original prompt tokens:}
\STATE $\mathbf{Z}=[\mathbf{e}_i, ..., \mathbf{e}_m]$ where $\mathbf{e}_i = \mathbf{E}(w_i)$, $\mathbf{e}_i \in \mathbb{R}^d$
\FOR{$1, ..., T$}
    \STATE {\color{gray} Random sample $b$ images from $\mathbf{I}$: } 
    \STATE $\mathbf{I}_b \subseteq \mathbf{I}$
    \STATE {\color{gray} Map continuous embeddings to discrete tokens:}
    \STATE $\mathbf{Z'}=E^{-1}_{\mathbf{E}}(\mathbf{Z})$
    \STATE {\color{gray}Calculate the gradient w.r.t. the \textit{discrete} embedding:}
    \STATE $g = \nabla_{\mathbf{P'}} \mathcal{L}(\mathbf{Z'}, \mathbf{I_b}, \theta)$
    \STATE {\color{gray} Apply the gradient on the \textit{continuous} embedding:}
    \STATE $\mathbf{Z} = \mathbf{Z} - \gamma g$
\ENDFOR
\STATE {\color{gray} Final inverted discrete homogeneous dimensions:}
\STATE $\mathbf{t_1} = E^{-1}_{\mathbf{E}}(\mathbf{Z})$
\STATE \textbf{return} $\mathbf{Z'}$
\end{algorithmic}
\end{algorithm}

To tackle this challenge, we propose a new method called \textit{prompt inversion}, which modifies the input prompt $\mathbf{t_0}$ to a new prompt $\mathbf{t_1}$ to capture shared features of the generated image set. Effectively, we are ``inverting'' the set of images into a caption describing the common dimensions. This method is more scalable than prompting large vision-language models in terms of both computational efficiency and cost, as it avoids repeated multimodal inference and can be generalized across image sets with minimal overhead. The resulting captions will reveal model-preferred homogeneous dimensions of $\mathbf{t_0}$.
% , and (2) acting as a filter to identify and remove overly similar/uniform images in later stages of generation
For example, given an initial prompt $t_0$ of \textit{``An ancient artist is composing a piece of work''}, it may be optimized into a prompt $t_1$ like \textit{``considering experienced beard apostle writing''}, which reveals implicit model preferences that would otherwise remain hidden.
In this case, $t_1$ explicitly reveals the model's tendency toward a bearded, meditating elderly male who is writing. 
% This optimized prompt $t_1$ presents implicit model preferences that may otherwise remain hidden and can be used to inform both diversification and filtering in subsequent stages.

Formally, prompt inversion starts by padding the original prompt $t_0$ with random tokens, and subsequently converting this padded prompt into a sequence of learnable, continuous token embeddings $\mathbf{Z}=[\mathbf{e_i}, ... \mathbf{e_m}], \mathbf{e_i} \in \mathbb{R}^{d} $, where $m$ is the number of homogeneous dimensions we seek to discover (a hyperparameter we set to 15 \revised{based on empirical findings to balance descriptiveness and coherence}) and $d=1024$ is the standard embedding dimension in CLIP.  
We assume access to a frozen CLIP model $\theta$~\citep{radford2021learningtransferablevisualmodels} that scores the similarity between text and images. 
In each iteration of prompt inversion, we randomly sample $b$ images from the entire set of images $\mathbf{I}$ and use the CLIP model to compute the prompt-image similarity between the $b$ images $\mathbf{I_b}$ and the $M$ current token embeddings, denoted as $\theta(\mathbf{Z}, \mathbf{I_b})$. This similarity will start very low, and our goal is to maximize this similarity so that token embeddings can faithfully capture the common homogeneous dimensions that are present but latent among the images. Towards this end, we define an objective function:
% \todo{in algorithm 1, calculate the grant is in the algorithm 1 but not in the text, projected embedding is not clear what projection means, how we apply gradient to continuous embedding is not explained in text.}
% \todo{$\mathcal{L}$ is not defined in algorithm 1. }
\begin{equation}
    \mathcal{L}(\mathbf{Z}, \mathbf{I_b}, \theta) = 1 - \theta(\mathbf{Z}, \mathbf{I_b})
\end{equation}
which maximizes the cosine similarity between two vectors.

The inverted token embeddings $\mathbf{Z}=[\mathbf{e_i}, ... \mathbf{e_m}]$ are $d$-dimensional continuous vectors and not interpretable by users, so we map these embeddings back to discrete tokens by finding the nearest neighbor token in the embedding matrix $E^{|V|\times d}$ through a projection function:

\begin{equation}
    E^{-1}_{\mathbf{E}}(\mathbf{e}) = \argmax_{j \in \{1, \ldots, |V|\}} \cos(\mathbf{e}, E_j) = \arg\max_{j} \frac{\langle \mathbf{e}, E_j \rangle}{\|\mathbf{e_i}\| \cdot \|E_j\|} 
\end{equation}
To ensure the resulting token be semantically related or easy to interpret, during forward pass, we map $\mathbf{Z}$ to its nearest discrete token embeddings $\mathbf{Z'}$ using projection $E^{-1}$: $\mathbf{Z'} = E^{-1}_{\mathbf{E}}(\mathbf{Z}):=[E^{-1}_{\mathbf{E}}(\mathbf{e_i}), ..., \\ E^{-1}_{\mathbf{E}}(\mathbf{e_m})]$. Then we compute the gradient w.r.t $\mathbf{Z'}$ using $g = \nabla_{\mathbf{Z'}} \mathcal{L}(\mathbf{Z'}, \\ \mathbf{I_b}, \theta)$ and update continuous embedding $\mathbf{Z}$ through $\mathbf{Z} = \mathbf{Z} - \gamma g$. 
This ensures that inversion remains grounded in the discrete token space at every step, leading to more interpretable and semantically meaningful prompts. The resulting interpretable homogeneous dimensions are denoted as $\mathbf{t_1} = E^{-1}_{\mathbf{E}}(\mathbf{Z})$. Empirically, we find that the inverted prompts $\mathbf{t_1}$ are human readable and semantically informative, even though including a few of gibberish words. Importantly, those meaningful words tend to capture core attributes shared across the image set. For example, as illustrated in Figure \ref{fig:pipeline}, object attributes like \textit{considering, experienced apostle with beard}, action like \textit{writing manuscript}, and thematic elements like \textit{prosperity character, bible} are successfully captured. \revised{We note that the few unreadable tokens do not degrade POET’s performance since LLMs largely ignore semantically unclear tokens during expansion. 
% In fact, they may contribute to interpretation of LLMs when sharing a similar embedding space.
} Algorithm \ref{alg:po} summarizes the steps involved in prompt inversion.
%with a projection function $\text{Proj}_{\mathbf{E}}$, that takes the individual embedding vectors $\mathbf{e_i}$ in the prompt and projects them to their nearest neighbor in the embedding matrix $E^{|V|\times d}$ where $|V|$ is the vocabulary size of the model. 

\subsection{Expansion of Homogeneous Dimensions}

Given these $m$ homogeneous dimensions identified in the first step, the second step is \textit{diversification}: automatically diversifying the set of generated images by prioritizing these dimensions to expand the input prompt. To achieve this, we prompt a large language model (e.g., \textit{GPT-4o}~\citep{openai2024gpt4ocard}) to sample attributes that modify the homogeneous dimensions. For instance, as illustrated in Figure \ref{fig:pipeline}, if the discovered homogeneous dimensions include \textit{``considering experienced beard apostle''} when prompted with \textit{``An ancient artist is composing a piece of work''}, diversified variants such as \textit{``a young male sculptor'', ``an elderly Asian female'' or ``a young Egyptian woman''} will be generated. Focusing on the expansion of discovered homogeneous dimensions ensures that no extra sampling is wasted on dimensions in which the model already exhibits diversity (e.g., whether the man wears a headscarf). To structure this expansion, we first categorize each homogeneous dimension into semantic groups (i.e., subjects, attributes, contextual settings, actions, and relationships) using the language model itself. Then, we instruct the language model to sample from those categories and construct replacements, ensuring that underrepresented concepts are prioritized. For each category, the homogeneous dimension will be refined or expanded into specific subsets (e.g., for action category, ``composing'' can be expanded to ``painting'' or ``chiseling''), and the system uniformly samples from the refined attributes while preserving the semantic consistency of the original prompt (The instruction we use for automatic expansions is located in Appendix Section \textit{Task Prompts}). Figure~\ref{fig:diversification} illustrates an example of our automatic diversification pipeline, highlighting how priority expansion is applied.

However, there is tension between output expansion and faithfulness to the original prompt - simply uniformly generating new attributes could potentially lead to over-expansion and images that no longer resemble the original semantic meaning of the user prompt. Therefore, we balance these two objectives via a filtering function
\begin{equation}
    \mathcal{F}(\hat{t}) =  \text{Div}(\hat{t})+ \lambda\cdot\text{Sim}_{\text{text}}(\hat{t}, t_0)  ,
\end{equation}
where $\hat{t}$ is a candidate expanded prompt and $\lambda$ is a tunable hyperparameter. Lower $\lambda$ encourages diversity, and higher $\lambda$ encourages consistency with the original prompt. Here, $\text{Sim}_{\text{text}}(\cdot)$ measures the cosine similarity between the candidate and the the original prompt in the CLIP text embedding space; $\text{Div}(\cdot)$ quantifies the diversity of of the image $I_{\hat{t}}$ generated from the candidate prompt $\hat{t}$:
\begin{equation}
    \text{Div}(\hat{t}) = \text{Sim}_{\text{image}}(t_0, I_{\hat{t}}) - \text{Sim}_{\text{image}}(t_1, I_{\hat{t}}),
\end{equation}
where $t_1$ is the inverted prompt that captures the homogeneous pattern observed in the original image set $\mathbf{I}$. Here $\text{Sim}_{\text{image}}$ denotes CLIP text-image similarity, lower $\text{Sim}_{\text{image}}(t_1, I_{\hat{t}})$ indicates $I_{\hat{t}}$ is more diverse. To balance faithfulness, we normalize it by adding similarity to the original prompt $\text{Sim}_{\text{image}}(t_0, I_{\hat{t}})$, ensuring that we only reward diversity if the resulting image still aligns with the original prompt.

\begin{figure}[t]
    \centering
    \vspace{-2mm}
    \includegraphics[width=1\linewidth]{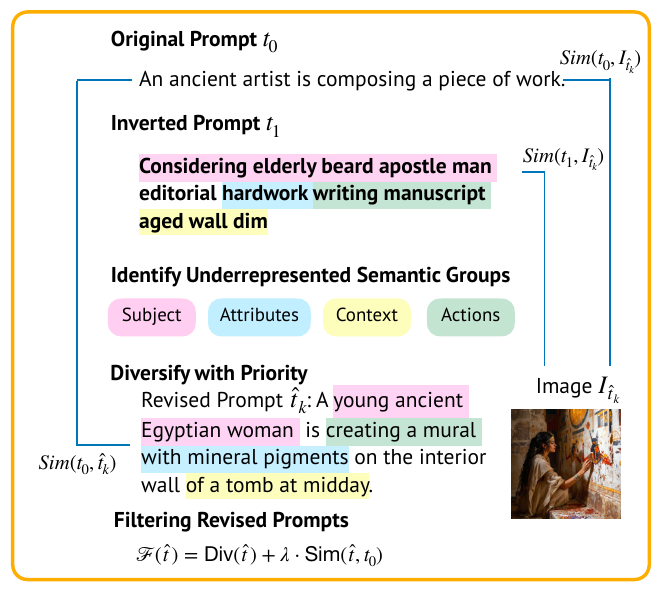}
    \caption{Automatic expansion of homogeneous dimensions. \model\ identifies and categorizes homogeneous concepts categorize $t_1$ into underrepresented semantic groups (highlighted). These groups are prioritized for expansions (e.g., $\hat{t_k}$, $I_{\hat{t_k}}$), followed by a filtering step that balances the diversity with semantic consistency to the original prompt $t_0$.}
    \label{fig:diversification}
\end{figure}

\subsection{Personalized Expansion with Feedback}

So far, in each step, the system discovers the common homogeneous dimensions through inversion, balances diverse prompt expansion with semantic consistency, and displays 
% the expanded prompts to the user alongside 
newly generated images for each. Here, we introduce the ability for \model~ to personalize to user preferences during re-prompting using feedback gathered from previous iterations of prompting.
Here, we assume two forms of user feedback before re-prompting (1) satisfaction rating on the previously generated image set, and (2) choices on a pair of images they most/least preferred from the set generated with expanded prompts (after diversification) as illustrated in Figure \ref{fig:pipeline}. \model~ learns from this interaction to enhance both the possibilities and efficiencies in the first two steps. 
As a result, the \model~system improves through the following actions:
\begin{itemize}[noitemsep,topsep=0pt,nosep,leftmargin=*,parsep=0pt,partopsep=0pt]
    \item Identifying which homogeneous dimensions the user cares the most about, among all possible bias dimensions detected through concept inversion, and prioritizing those dimensions for faster personalization.
    \item Identifying which ways of expanding the prompt the user cares the most about and prioritizing those possible trajectories for faster personalization.
\end{itemize}

Learning is performed by \textbf{contextual adaptation via preference conditioning}. ~\citet{chen2024tailoredvisionsenhancingtexttoimage} finds that historical user interactions can be leveraged to enhance user prompts via analyzing high-frequent words. We go further by storing users' past feedback--both their re-prompts and satisfaction scores--in a user \textit{preference profile}. During future prompt expansions, this profile will be appended as a conditioning context for the large language model. Specifically, we prompt the large language model to (i) identify which homogeneous dimensions or categories (e.g., subject attributes, contextual settings, etc.) the user consistently focuses on, and (ii) preserve patterns from prior prompt revisions that correlate with higher satisfaction, while avoiding or adjusting features associated with lower ratings. Additionally, the user’s most/least preferred images are analyzed via a multimodal large language model (e.g., GPT-4o in our case) to extract visual and semantic patterns that the user prefers or tends to avoid (See Appendix Section \textit{Task Prompts} for our prompt instruction). This includes attributes in the image such as age, ethnicity, context setting, composition, tone, among others. These patterns are then summarized and incorporated into the user's profile as additional signals to guide future prompt refinement. This preference-driven contextual adaptation allows the model to dynamically adapt its expansion strategy to individual users without fine-tuning model weights.

\subsection{Implementation Details}

To summarize, our full model \modelfour\ combines the power of identifying and expanding homogeneous dimensions generated by text-to-image models and personalized learning from user feedback on expanded dimensions. To test the specific components of our method, we also implement and compare to \modelone\ which is a pre-trained text-to-image generative model Flux.1-dev (or any open-weight T2I model), \modeltwo\ that identifies and expands common dimensions of homogeneity across but does not perform personalized learning, and \modelthree\ which personalizes a base Flux.1-dev model based on users' prompt and image preferences but does not automatically identify and expand homogeneous dimensions.

In this section, we provide implementation details for each of these models for reproducibility. For \modelone, we set the guidance scale to 7.5 and the number of inference steps to 28. For each prompt, we generate 10 different images with random seeds. For \modeltwo, we use the open-source CLIP model, OpenCLIP-ViT/H~\citep{Cherti_2023} as the backbone. During the prompt inversion process, we use a learning rate of 0.1 and run 1000 optimization steps using the AdamW optimizer~\citep{loshchilov2019decoupledweightdecayregularization}.  The batch size $b$ for the sampled image set is as 2 and the embedding dimension $d$ is fixed to 1024 (as in CLIP), $\lambda$ is set to 0.1. After expansion, we select 10 prompts. Experiments for automatic evaluations were conducted using a single H100 GPU with 80GB memory. Inference takes around 0.5 seconds per text prompt. User studies with \modelthree\ and \modelfour\ are deployed in Huggingface Spaces, using a single L40S GPU with 62GB memory.

% - appending prior user preferences as context into the model,
% \begin{enumerate}
% [noitemsep,topsep=0pt,nosep,leftmargin=*,parsep=0pt,partopsep=0pt]
%     \item \textbf{Contextual adaptation via preference conditioning:} the user's past feedback--both their re-prompts and satisfaction scores--is stored in a user preference profile. During future prompt expansions, this profile is appended as conditioning context to the \textsc{Prompt Refiner}. Specifically, we prompt the large language model to (i) identify which bias dimensions or categories (e.g., subject attributes, contextual settings, etc.) the user consistently focuses on, and (ii) preserve patterns from prior prompt revisions that correlate with higher satisfaction, while avoiding or adjusting features associated with lower ratings. 

% In addition, the user’s most liked and disliked images are analyzed to extract visual and semantic patterns that the user prefers or tends to avoid. This includes attributes such as age, ethnicity, setting, composition, tone and more—automatically extracted via vision-language models (GPT-4 in our case). These patterns are then summarized and incorporated into the user's profile as additional signals to guide future prompt refinement. This preference-driven contextual adaptation allows the model to dynamically adapt its expansion strategy to individual users without fine-tuning model weights.

% - updating the text-to-image model weights with respect to selected images,
% \item \textbf{Fine-tuning text-to-image model with preference data:}
% - \todo

% \end{enumerate}
\section{Automatic Evaluations}

Our first set of evaluations involve automatic quantitative metrics to summarize the accuracy of each step in \model. Specifically, we seek to measure how accurate our method is in identifying homogeneous dimensions as well as the creativity and consistency trade-offs in expanding homogeneous dimensions.

\subsection{Setup}

We use the image-captioning dataset Flickr30K~\citep{plummer2016flickr30kentitiescollectingregiontophrase} for automatic evaluation, randomly sampling 1,000 text prompts to generate corresponding images. We use the open-weights pre-trained text-to-image model Flux.1-dev~\citep{flux2024} as the backbone generative model for all experiments. For each selected prompt, we generate 10 images from all baseline models and variations of our \model\ method.

\subsection{Results}

\subsubsection{Evaluating identification of homogeneous dimensions}

\begin{table}[t]
\centering
% \resizebox{0.\linewidth}{!}{%
\begin{tabular}{lc}
\toprule
\textbf{Homogeneous Dimension Identification} & ICAD ($\uparrow$) \\
\midrule
\textit{Open-sourced models} & \\
BLIP-2+LLaMA-7b & 0.425\\
LLaVA-1.5-7b & 0.437 \\
\hdashline\noalign{\vskip 1ex}
\textit{Close-sourced models} & \\
GPT-4o & 0.470 \\
Ours (Prompt Inversion) & 0.482 \\
\bottomrule
\end{tabular}
% }
\caption{ICAD scores of images generated by our pipeline with different homogeneous dimension identification strategies. Prompt inversion yields the highest diversity among all methods.}
\vspace{-7mm}
\label{tab:ablation-strategy}
\end{table}

To evaluate the effectiveness of our strategies for identifying homogeneous dimensions, we compare it with three alternative methods that also strive to identify common homogeneous dimensions among a set of generated images: 
\begin{itemize}
[noitemsep,topsep=0pt,nosep,leftmargin=*,parsep=0pt,partopsep=0pt]
    \item Directly prompting a pre-trained vision-language model like LLaVA-1.5-7b (open-source) or GPT-4o (closed-source) to analyze and summarize the common homogeneous dimensions across the 10 images.
    
    \item Using a pre-trained vision-language model like BLIP-2 to generate individual captions for each of the 10 images, followed by a large language model (e.g., LLaMA2-7b) to summarize the common dimensions across these 10 captions.
\end{itemize}

To ensure a fair comparison, we replace the prompt inversion algorithm used in our system with different strategies as outlined above, keeping all other components of \modeltwo\ consistent. We evaluate each method in terms of expanded image diversity and inference cost. 

As presented in Table \ref{tab:ablation-strategy}, both pre-trained vision-language models BLIP-2 and LLaVA-1.5-7B struggled to accurately summarize shared visual attributes such as age and background across the full image set. Empirically, we observe that BLIP-2 often extracts attributes from individual images without comparing its occurrence across all images in the set. When passed to a language model for summarization, noisy or non-representative details also tend to propagate into the final dimensions (e.g., see Appendix for a detailed comparison; we show an example that summarizes images to ``\textit{an Asian woman sitting...}'' when only one image depicts an Asian woman). These models are primarily optimized for single-image input and have not been fine-tuned on multi-image inputs, so they typically generate incomplete summaries based only on a few images. Alternatively, more capable multi-image models, like GPT-4o, can better handle joint visual reasoning over multiple images \revised{(still fall behind POET, 0.470 vs. 0.482)}, but are closed-sourced and costly to access.  Identifying homogeneous dimensions from 10 images per text prompt takes approximately 3 seconds and costs around \$0.05, making them impractical for large-scale automated analysis. 
% \revised{Our primary objective is to improve overall system-level performance. Systematically, {\model} significantly outperforms GPT-Image-1. While GPT-4o indeed achieves relatively good local HDI metrics in some cases, ICAD score for local HDI only reflect how well a method identifies homogeneous dimensions—this does not directly correspond to improved system-level image generation quality. This highlights the importance of our full pipeline.}
% We agree that reducing reliance on closed-source models is desirable and will explore smaller open-source models in future work.}

\subsubsection{Evaluating diversity of dimension expansion}

To evaluate the effectiveness of our dimension expansion strategy, we use the Intra-Class Average Distance (ICAD) metric~\citep{le18_interspeech}, which quantifies the average pairwise diversity of images generated from the same prompt. ICAD values range from 0 to 1, with higher scores indicating greater visual diversity and lower scores reflecting more homogeneity within image sets.

% \begin{figure}[t]
%     \centering
%     \begin{subfigure}[b]{0.49\columnwidth}
%         \centering
%         \includegraphics[width=\linewidth]{figures/charts/auto_icad.png}
%         \caption{}
%         \label{fig:image1}
%     \end{subfigure}
%     \hfill
%     \begin{subfigure}[b]{0.49\columnwidth}
%         \centering
%         \includegraphics[width=\linewidth]{figures/charts/auto_icad_gpt.png}
%         \caption{}
%         \label{fig:image2}
%     \end{subfigure}
%     \caption{Evaluation of image diversity with ICAD (the average pairwise diversity of image set). HDI denotes homogeneous dimension identification. (a) {\modeltwo} consistently generates more diverse images than {\modelone}. HDI further raises image diversity. (b) Comparison with GPT-4o.}
%     \label{fig:auto-eval}
% \end{figure}

\begin{figure}[t]
\centering
{\includegraphics[width=0.8\linewidth]{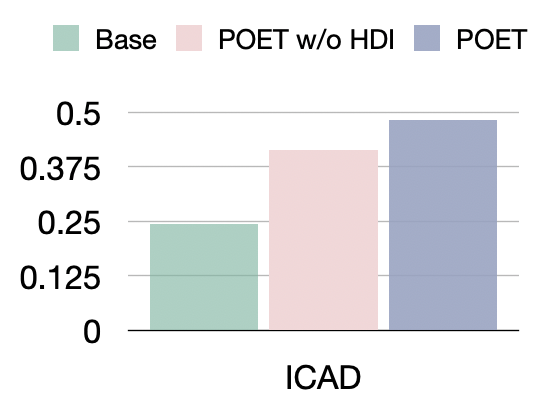}}%
\vspace{-2mm}
\caption{Evaluation of image diversity with ICAD (the average pairwise diversity of image set). HDI denotes homogeneous dimension identification. {\modeltwo} consistently generates more diverse images than {\modelone}. HDI further raises image diversity. }
\label{fig:auto-eval}
\end{figure}

\begin{figure}[t]
\centering
{\includegraphics[width=0.8\linewidth]{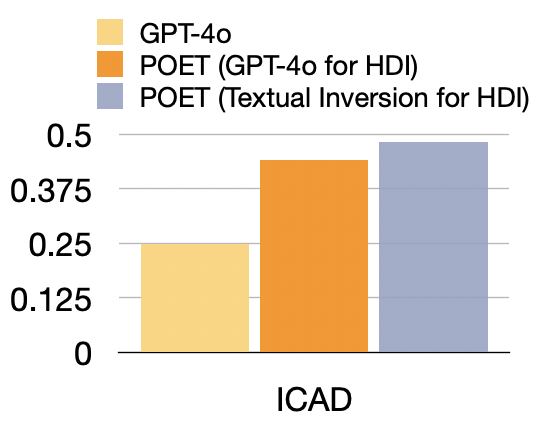}}%
\vspace{-2mm}
\caption{Comparison with GPT-4o}
\label{fig:auto-eval-gpt}
\end{figure}

\begin{figure*}[t]
\centering
{\includegraphics[width=\textwidth]{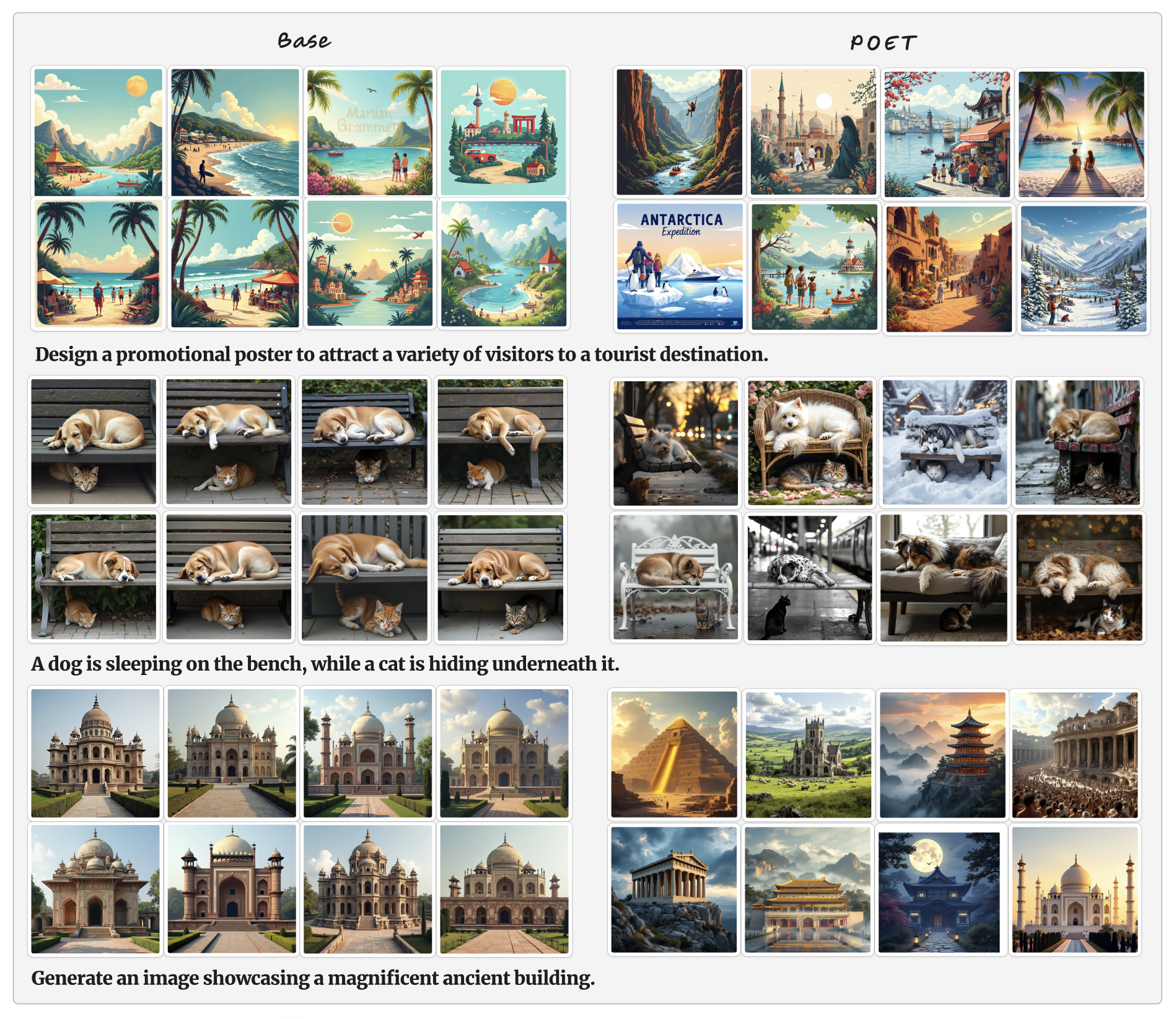}}%
\caption{Diversity comparison to \modelone\ model. The \modelone\ exhibits reduced diversity, while \modeltwo\ consistently maintains diversity in subject and context across different prompts. For example, tourist destinations vary widely, dog breeds and backgrounds are diverse, and the styles of ancient buildings differ significantly.}
\label{fig:qual-diverse}
\end{figure*}

As shown in Figure~\ref{fig:auto-eval} (a), the baseline {\modelone} model produces relatively low ICAD scores, suggesting limited variation across its generated outputs. When we apply prompt expansion without first identifying homogeneous dimensions ({\modeltwo} w/o HDI), we observe a substantial improvement in diversity from 0.24 to 0.41. However, by incorporating homogeneous dimension identification (HDI) through prompt inversion and filtering, \modeltwo\ achieves the highest ICAD score—rising from 0.41 to 0.48—indicating the model effectively filters similar images that share homogeneous dimensions as images generated by \modelone.  We also display qualitative results comparing {\modelone} and {\modeltwo} in Figure \ref{fig:qual-diverse}. 

\revised{To compare POET’s performance with closed-source GPT-4o, we compare the following three systems:
\begin{itemize}
    \item GPT-4o for image generation, with GPT-Image-1 as backend. 
    \item {\model} pipeline with GPT-4o serving as the HDI strategy. 
    \item {\model}, using prompt inversion as the HDI strategy.
\end{itemize}
As shown in Figure~\ref{fig:auto-eval-gpt}, GPT-4o alone produces highly similar images (ICAD = 0.248), while \model\ achieves a substantially higher ICAD score of 0.481. When substituting GPT-4o for prompt inversion as the HDI component in our pipeline, we observe a slight drop in performance—but the system still generates significantly more diverse images than using GPT-4o alone. This highlights an important point: the improvement is not due to GPT-4o itself, but rather the design of our pipeline, which remains central even when different HDI strategies are used. Moreover, using GPT-4o as the HDI module introduces considerable overhead in the number of required queries, making it both time- and cost-intensive. In contrast, our full POET system, with prompt inversion, is not only more effective but also significantly more efficient. These results demonstrate that even powerful models like GPT-4o cannot substitute for our tailored approach—our pipeline is what drives both diversity and efficiency.}

\subsubsection{Evaluating consistency of dimension expansion}

\begin{figure}[t]
    \centering
    \vspace{-2mm}
    \includegraphics[width=\linewidth]{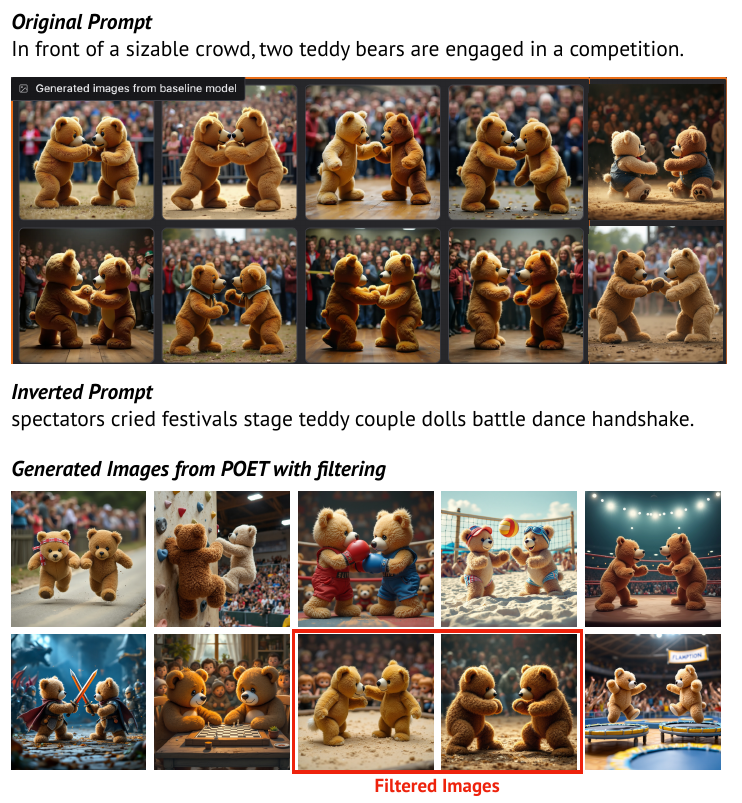}
    \caption{Images generated from \modeltwo\ are diverse but consistent to the original prompt. The filtered images using homogeneous dimensions are similar to the original image set. }
    \label{fig:visualization}
\end{figure}

There can be tension between diversity and faithfulness - simply uniformly generating new attributes could potentially lead to over-diversity and images that no longer resemble the original semantic meaning of the user prompt. To evaluate the consistency of expanded prompts with original user intents, we compare the images generated after expansion with the original prompt in Figure \ref{fig:visualization}. Given the initial prompt \textit{``In front of a sizable crowd, two teddy bears are engaged in a competition.''}, the generated images exhibit strong visual similarity. {\model} successfully identified homogeneous dimensions such as a stage-like background, two teddy bear characters in a confrontational pose, engaging in actions resembling dancing or handshaking. {\model} successfully identifies these homogeneous dimensions. After diversification, the new images depict two teddy bears competing in different ways—e.g., racing, playing chess—while remaining faithful to the core semantics of the prompt. Furthermore, {\model} also filters out expanded prompts that produce images nearly identical to the original generations, demonstrating its ability to filter and avoid redundancy based on identified homogeneous dimensions.

% \subsection{Findings}
% \todo automatic evaulation findings

\begin{table*}[h]
\centering
\resizebox{\textwidth}{!}{%
\begin{tabular}{cp{0.5\linewidth}p{0.25\linewidth}}
\toprule
\textbf{Use Case} & \textbf{Background \& Instructions to Participants}  & \textbf{Initial Prompts} \\
\hline
(S1) Product Advertisement & You are designing an advertising campaign for a new line of coffee machines. To ensure the campaign resonates with a wider audience, you use generative models to create marketing images that showcase a variety of users interacting with the product. & Design an advertisement image showcasing a range of users operating coffee machines. \\
\\
(S2) Tourist Promotion & You are creating a travel campaign to attract a variety of visitors to a specific destination. To make the promotional materials more engaging, you use generative models to design posters that highlight a broader array of experiences. & Design a promotional poster to attract a variety of visitors to a tourist destination.\\
\\
(S3) Fictional Character Generation & You are creating a superhero video game that’s fun and relatable to a range of users. You decide to use generative models to help visualize a new character. & Design a video game superhero character that is relatable.\\
\\
(S4) Interior Design & You are helping design the furniture layout for a model one-bedroom rental apartment. To make the apartment appealing to different potential tenants, you try to visualize different furniture placements before setting everything up. & Design an interior of an apartment that’s appealing to potential tenants.\\
\bottomrule
\end{tabular}
}
\caption{Creative use cases for generative models: a set of design challenges where each scenario provides a basis for participants to explore text-to-image generation. }
% \todo{explain a bit more and make caption more exciting}}
\label{tab:scenarios}
\end{table*}

\section{User Study}

To understand how users perceive and interact with different versions of the \model~ system, we conducted a within-subjects study where each participant evaluated the four versions of the \model~ system across three phases. In the following section, we first introduce the experimental setup (\S\ref{exp_design}), followed by our recruitment and analysis methods (\S\ref{recruit}). Finally, we present quantitative results along with qualitative think-aloud findings (\S\ref{user_findings}).

Our study was guided by the following \textbf{research questions}: 
% \begin{itemize}
\begin{enumerate}
    \item[\textbf{RQ 1}] Can \modeltwo\ detect and expand homogeneous dimensions in text-to-image models to produce outputs that users perceive as more visually diverse than outputs of the \modelone\ model?
    \item[\textbf{RQ 2}] Does \modeltwo\ produce image outputs that help users achieve creative results that are more desirable or that take fewer prompt iterations than images generated by the \modelone~ model?
    \item[\textbf{RQ 3}] How does \modelfour\ help users generate more preferred creative outputs, or do so in more efficient ways than \modelthree?
\end{enumerate}

\subsection{Experimental Design}\label{exp_design}
For all three phases of the experiment, we constructed scenarios representing ideation processes of four distinctive creative domains: product advertisement, tourist promotion, fictional character generation, and interior design -- Table \ref{tab:scenarios} details full instructions. Participants engaged with the system through three phases. Across phases, we requested and reminded participants to think aloud, so we may gather their (1) reactions to the interface and generated images, as well as (2) rationales and thought processes as they deliberated on Likert-scale ratings and choices.
We evaluated this design with three pilot participants to improve user experience with the system and adjust task timings. 
Below, we explain the assigned tasks of each phase.

\subsubsection{Phase 1: Measuring Improvements in Homogeneous Dimensions -- Evaluating Image Difference of \modelone\ v.s. \modeltwo}
In the first phase, we sought to evaluate whether users perceived image groups generated by the \modeltwo\ system as more varied than the \modelone\ baseline (the plain pre-trained model flux.1-dev) where no expansion of homogeneous dimensions is performed before prompting for images. Each participant was presented six pairs of images for each of the four scenarios summarized in Table \ref{tab:scenarios} (for a total of 24 pairs). 
We introduced participants to each scenario by presenting textual descriptions of the scenario (labeled as Background), followed by the prompt used to generate the pair.
For each pair, images were either (a) both generated using \modelone\ or (b) both generated with the expanded \modeltwo~ backend, and participants were tasked with rating how different the two images are from each other on a Likert 7-point scale (1 = ``Very Similar'', 7 = ``Very Different''). In each scenario, three of the six pairs were generated with \modelone, while the other half consisted of \modeltwo\ outputs. We counterbalanced the order of scenarios between participants so that each possible permutation of scenario ordering (a total of 4! = 24) was followed by at least one participant (see Appendix for the complete list of ordering assignments between participants). 

\subsubsection{Phase 2: Prompting with Expansions -- Prompting \modelone\ v.s. \modeltwo}

% comparing \modelone\ with \modeltwo

After evaluating pairwise image similarity across all scenarios, participants proceeded to Phase 2, where they iteratively prompted two underlying models for one of the four creative scenarios. 
To perform within-subject comparisons, we requested that each participant worked with the same scenario twice to expose them to both experimental (\modeltwo~) and baseline (\modelone~) conditions. We assigned scenarios in a way where every scenario was seen by 6 participants and counterbalanced the order of conditions (experimental \& baseline) to reduce learning effects on our results -- a full list of scenario assignments and conditions is in the Appendix.

% After completing this process for one system condition, participants repeated the entire task with the condition they had not yet experienced (i.e., the baseline if they had first completed the experimental condition, and vice versa). 
% This design 

First, the interface re-introduced the selected scenario using the same background descriptions found in Table~\ref{tab:scenarios}, along with an initial pre-written prompt (found in the right-most column of Table \ref{tab:scenarios}) and a corresponding set of four images generated from this prompt. 
% Each participant was shown 
This initial image set for each scenario was fixed across participants to ensure a consistent starting point for the prompting task. Participants started by rating their satisfaction with the initial image set on a 7-point Likert scale (1 = ``Very Unsatisfied'', 7 = ``Very Satisfied''), followed by at least one iteration of prompting (where participants could iterate on the provided prompt or their own from scratch) after editing the prompt in some way --- we allowed for a maximum of 5 re-prompts to stay within the time limits of our study. In the subsequent iteration(s), participants also rated their satisfaction with the image sets generated from each new prompt and were allowed to continue prompting or stop when they selected ``Satisfied'' or ``Very Satisfied''. 
% If continuing, participants could iterate on the provided prompt or their own from scratch. 
No word limit was imposed on prompt entries. When participants stopped iterating on the prompt (after reaching satisfaction or 5 rounds), they selected their favorite image from among all generated outputs, and rated their satisfaction on a continuous scale from 1 to 10. 

% \begin{figure}[t]
%     \centering
%     \includegraphics[width=1\linewidth]{figures/interface-phase3.png}
%     \caption{Interface - Phase 3.During each round of prompting, participants were asked to identify their most satisfying image and their least satisfying image among all images generated up to that point. }
%     \label{fig:phase3}
% \end{figure}

\begin{figure*}
    \centering
    \vspace{-2mm}
    \includegraphics[width=1\textwidth]{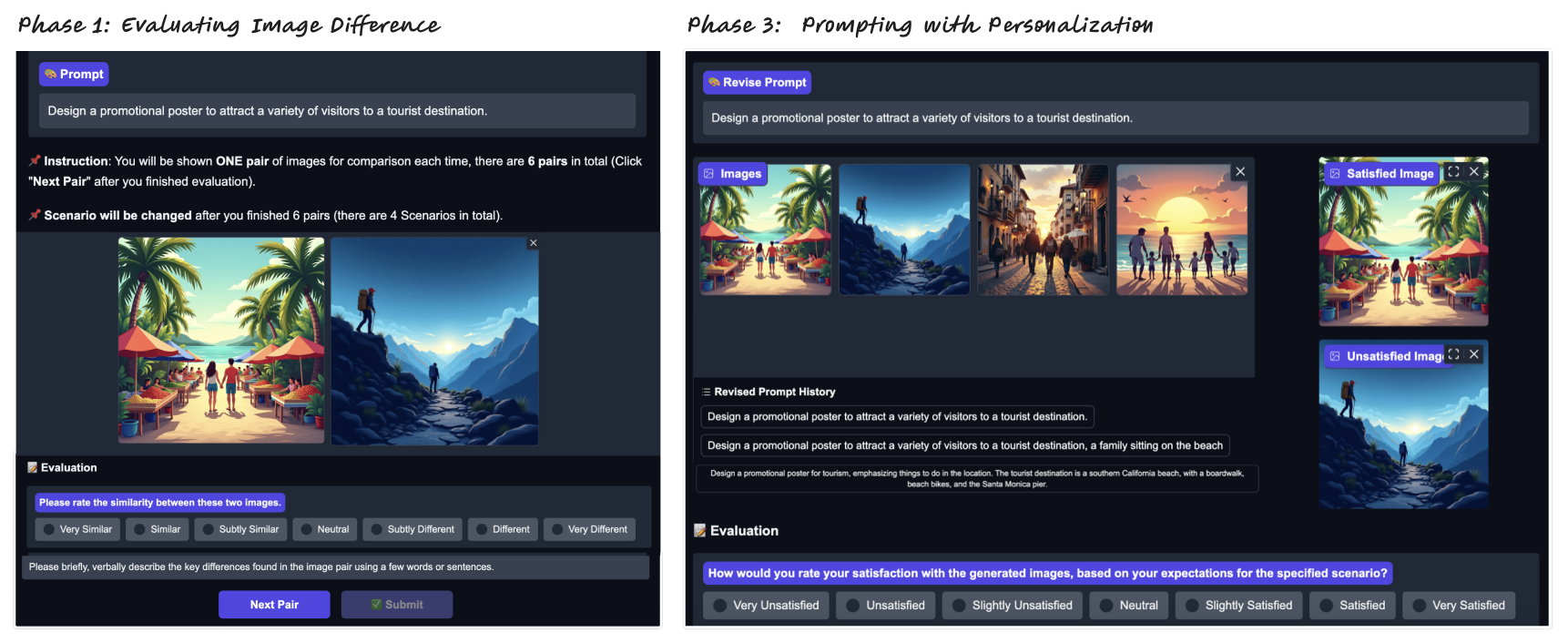}
    \caption{Interfaces -  Phase 1: rating image differences (left). Phase 3: During each round of prompting, participants were asked to identify their most satisfying image and their least satisfying image among all images generated up to that point (right).}
    \label{fig:interface-big}
    \vspace{-3mm}
\end{figure*}

\subsubsection{Phase 3: Prompting with Personalization -- Prompting with \modelthree\ \& \modelfour}

Phase 3 followed the same procedure as Phase 2, but focused on evaluating participant perceptions of personalization using the baseline \modelthree~ and the experimental \modelfour~ models. The conditions are set up in a similar manner as Phase 2 to perform within-subject comparison on effects of personalization, but each participant worked with a new scenario that's different from that of Phase 2.
During each round of prompting, participants were asked to identify their \textit{most satisfying} image (i.e., favorite) and their \textit{least satisfying} image (i.e., least favorite) among all images generated up to that point 
--- these selections were cumulative, with participants evaluating their preferences across the entire set of images generated throughout the session, rather than only the most recent iteration of prompting. 

Participants filled out a questionnaire that asked them to optionally disclose (1) demographic information, (2) experience with everyday discrimination, and (3) familiarity with AI-powered systems. We based our questions about everyday discrimination on the “Everyday Discrimination Scale” \cite{williams1997racial, carter2017prevalence, harnois2019measuring}, a widely used instrument that measures how frequently individuals experience perceived discrimination in daily life. This scale has been used in prior research to examine how individuals’ experiences with discrimination may influence their interactions with AI systems~\cite{kingsley2024investigating}.
\subsection{Recruitment \& Analysis} \label{recruit}

\paragraph{Recruitment}
After obtaining IRB approval for our study from the leading institution, we recruited participants from college campuses in the United States through a combination of digital outreach methods, including Slack workspaces and courses. Potential participants were asked to fill out a form indicating their area of study, from which we oversampled participants with familiarity with AI systems and those studying in creative domains. Recruitment materials described the study's purpose, time commitment, and compensation details, in addition to linking to an IRB-approved consent form. Participation was voluntary, and informed consent was obtained before beginning the study. Participants were compensated with \$30 for sessions that lasted, on average, 60 minutes.

\paragraph{Thematic Analysis}
We conducted a thematic analysis of the think-aloud data to identify patterns in participants' reactions, decision-making processes, and interactions with the system. Following Braun and Clarke \cite{clarke2013thematic}, we first familiarized ourselves with the transcripts by reading them fully, then generated initial codes that captured notable behaviors and reflections. These codes were iteratively grouped into broader themes that reflected recurring user goals, frustrations, and strategies. Themes were reviewed and refined collaboratively among the research team to ensure they accurately represented the dataset and were relevant to our research questions. Throughout the analysis, we focused on how participants reasoned about prompt modifications, interpreted image diversity, and expressed satisfaction or dissatisfaction with the results. This qualitative analysis complemented our quantitative findings by revealing how and why participants arrived at their choices, offering deeper insight into their experiences with each condition.

\paragraph{Quantitative Analysis} 
% In phase 1 (Figure \ref{fig:phase1}), we collected participants' ratings of perceived visual difference for all pairs of images presented to them. In phases 2 and 3 (Figures \ref{fig:phase2} \& \ref{fig:phase3}), we asked participants to rate their sanctification with each set of generated images, including output of the initial provided prompt.

To examine whether such outcome variables --- perceived image difference in Phase 1 (See Figure \ref{fig:interface-big}, full interface images across three phases are located in Section 4 of the Appendix.), as well as degree of satisfaction and number of rounds required to achieve satisfaction in Phases 2 and 3 --- varied across our experimental conditions, we used a fixed-effects panel linear model for each phase of the study. In phase 1, we also report how each of the four scenarios interacted with the outcome of image difference.

\subsubsection{Positionality}
We recognize and reflect on how our interpretations of user study results are influenced by our own backgrounds, identities, and trainings as researchers, so as to avoid unintentionally substituting our own voices over our participants and target user populations.  
\revised{Our team consists of research from three US-based institutions, where we work on or receive training in the areas of Human-Computer Interaction, Machine Learning and Software Engineering. While several team members teach and employ design concepts in their work, none identify as visual artists. As such, our backgrounds may lead us to assess and define homogeneities and ``normativity'' of visual content in ways that may deviate from certain creative professionals.}
% [identifying reflections and positions are redacted for anonymity]

\subsection{Findings}
\label{user_findings}

% Research questions: 
% \begin{enumerate}
% \item[\textbf{RQ 1}] can we automatically detect and expand homogeneous dimensions of current text-to-image model outputs to increase perceivable visual variation?
% \end{enumerate}

% \begin{hypothesis}
%     The  found in images produced by  can be automatically detected and expanded upon to generate results with more visual variation.
% \end{hypothesis}
% \begin{hypothesis}
%     Images with reduced homogeneity along automatically-detected dimensions --- and thus more visual variation (H1) --- will help users reach more desirable outputs in creative tasks, in fewer iterations of prompting.
% \end{hypothesis}
% \begin{hypothesis}
% Personalization, especially when combined with expanded output spaces, will further help users generate more desirable images in creative tasks, in fewer prompting iterations.
% \end{hypothesis}
\subsubsection{Observations during Evaluations of Image Difference (Phase 1)}

\begin{figure}[t]
\vspace{-4mm}
\centering
% \subcaptionbox{Method}{\includegraphics[width=0.50\textwidth]{figures/charts/diversity_boxplot.png}}%
% \hfill
% \subcaptionbox{Method * Scenario}{
\includegraphics[width=0.50\textwidth]{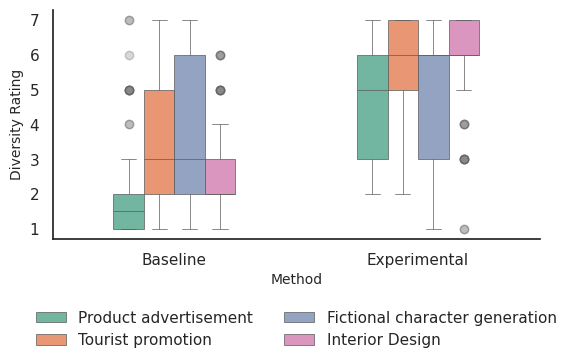}
% }%
\caption{Phase 1-Image Difference Rating. Experimental ({\modeltwo}) is rated as more different than the Baseline ({\modelone}) across all scenarios, with statistically significant differences. }
\label{fig:phase1}
\end{figure}

Figure \ref{fig:phase1} and Table \ref{tab:phase1} show that participants rated the image pairs of \modeltwo\ as more different than those from \modelone\ across all scenarios. The effect is more prominent for the fictional character generation (S3) and interior (S4) scenarios, where average participant ratings were higher for images of the experimental condition by 2.28 and 0.714 points on a 7-point Likert scale, respectively. The complete set of images used to construct pairwise comparisons is located in Section 4 of the Appendix.

\begin{table}[t]
\centering
\resizebox{\linewidth}{!}{%
\begin{tabular}{llc}
\toprule
\begin{tabular}[c]{@{}l@{}}Dependent Variable:\\ \textbf{Image Difference}\end{tabular}  & Coefficient & \begin{tabular}[c]{@{}c@{}}Standard\\        Error\end{tabular}        \\ \hline
\textit{\modeltwo\ vs \modelone}            & 2.28***  & 0.203        \\ \hline
\textit{\modeltwo\ vs \modelone*Interior Design}                    & 0.714**   & 0.257        \\ \hline
\textit{\modeltwo\ vs \modelone*Product Advertisement}              & -0.369    & 0.257        \\ \hline
\textit{\modeltwo\ vs \modelone*Tourist Promotion}                  & 0.357     & 0.257        \\ \hline
\multicolumn{1}{c}{Number of Observations}      & \multicolumn{2}{c}{672}      \\ \hline
\multicolumn{3}{c}{\begin{tabular}[c]{@{}c@{}}*** signifies a p-value \textless{}.001, ** denotes a p-value \textless{}.01\\ \hline
\textit{Reference scenario was set to Fictional Character Generation} 
\\ \hline
\textit{Difference was measured on a 7-point Likert scale where 1 is most similar}
\\ 
\end{tabular}} \\ \bottomrule
\end{tabular}
}
\vspace{1mm}
\caption{Phase 1 regression model predicting image difference ratings between \modelone\ \& \modeltwo. Participants rated the image pairs of \modeltwo\ as more different than those from \modelone\ across all scenarios; the effect is more prominent for the fictional character generation (S3) and interior (S4) scenarios, where average participant ratings were higher by 2.28 and 0.714 points on a 7-point Likert scale, respectively.}
\label{tab:phase1}
\end{table}

% add new line and standard error with coefficient

During Phase 1 of the think-aloud, where participants verbally described rationales for the ratings, we observed that selective attention played a large role in shaping some participants' perceptions of diversity and thus their ratings of differences in images in both conditions. In the tourist promotion scenario (S2), for example, participants tended to zero in on prominent visual features—such as terrain—which overshadowed broader contextual elements (e.g., tourist activities). 
Such ``selective attention'' sometimes caused higher diversity scores for the baseline outputs. For example, in the product advertisement scenario (S1), participants focused on demographic markers like gender, race, or age, while in the interior design task (S4), they focused on lighting. In both cases, they made quick decisions about a pair based on only a few selective visual elements, without accounting for more nuanced dimensions such as artistic style or engaged activities.

In contrast, the character generation task (S3) highlighted how individual interpretations of distinctiveness varied widely among participants. Some, like P1, P3, and P9, concentrated on easily identifiable demographic markers such as species or gender, while others picked up on subtler factors such as emotions and contextual elements -- for example, P3 perceived one character as ``angry'', while P5 was one of only two participants to notice the absurdity of using a coffee machine during a picnic. This divergence underscores a broader pattern: some participants relied on obvious visual markers, while others appreciated more nuanced details like emotional expression, pose, or stylistic differences. A similar divergence emerged in the interior design scenario (S4), where differences in ratings often reflected participants’ visual priorities. For instance, P22 emphasized furniture layout over background aesthetics, which led them to perceive less diversity in the experimental outputs.

% \begin{table}[]
% \resizebox{\linewidth}{!}{%
% \begin{tabular}{lll}
% \textbf{Dependent Variable} & \textbf{Rounds}        & \textbf{Satisfaction}   
% \\ \hline 
% & \begin{tabular}[c]{@{}l@{}}Coefficient\\ (Std Err)\end{tabular} & \begin{tabular}[c]{@{}l@{}}Coefficient\\ (Std Err)\end{tabular} 
% \\ \hline
% POET-Expand vs POET-Base  & -0.857.  & 0.600** \\ \hline
% \multicolumn{1}{c}{Number of Observations} & 56 & 235   \\ \hline
% \multicolumn{1}{c}{Number of Groups}   & TODO  &  \\
% \multicolumn{3}{c}{\begin{tabular}[c]{@{}c@{}}** signifies a p-value \textless{}.01, . denotes a p-value \textless{}.1\\ \hline
% % \textit{Reference scenario was set to Fictional Character Generation} 
% % \\ \hline
% \textit{Ratings were measured on a Likert scale of 7 (1 is most satisfied)}
% \\ 
% \end{tabular}} & \hline
% \end{tabular}
% }
% \caption{Phase 2 Analysis}
% \end{table}

\begin{table}[t!]
\centering
\resizebox{0.45\textwidth}{!}{%
\begin{tabular}{ccc}
\toprule
\textbf{Dependent Variables} & \textbf{Rounds} & \textbf{Satisfaction} \\ \hline
                             & Coefficient (Std Err) & Coefficient (Std Err) \\ \hline
\textit{\begin{tabular}[c]{@{}l@{}}\modeltwo\ vs \modelone\ \end{tabular}} & -0.857 . (0.453) & 0.600 ** (0.229) \\ \hline
No. of Observations & 28 & 28 \\ \hline
No. of Groups       & 56 & 235 \\
\multicolumn{3}{c}{. signifies a p-value \textless{}.1 and ** denotes a p-value \textless{} 0.01}    \\ \hline
\multicolumn{3}{c}{\textit{Satisfaction was measured on a 7 point Likert scale (1 is most unsatisfied)}}
\\ \bottomrule
\end{tabular}%
}
\caption{Phase 2 regression predicting iteration rounds and satisfaction rating between \modelone\ \& \modeltwo. Our regression results reveal that \modeltwo\ was able to help users achieve satisfiable results in fewer rounds (0.85 fewer on average) than \modelone, and generated significantly more satisfiable images overall (higher by 0.6 on a 7-point Likert scale). }
\vspace{-3mm}
\label{tab:phase2}
\end{table}

\begin{table}[]
\centering
\resizebox{0.45\textwidth}{!}{%
\begin{tabular}{ccc}
\toprule
\textbf{Dependent Variables} & \textbf{Rounds} & \textbf{Satisfaction} \\ \hline
                             & Coefficient (Std Err) & Coefficient (Std Err) \\ \hline
\textit{\begin{tabular}[c]{@{}l@{}}\modelfour\ vs \\ \modelthree\ \end{tabular}} & -1.35 ** (0.409) & 0.652 ** (0.217) \\ \hline
No. of Observations & 28 & 28 \\ \hline
No. of Groups       & 56 & 207 \\
\multicolumn{3}{c}{** signifies a p-value \textless{} 0.01}    \\ \hline
\multicolumn{3}{c}{\textit{Satisfaction was measured on a 7 point Likert scale (1 is most unsatisfied)}}
\\ \bottomrule
\end{tabular}%
}
\vspace{1mm}
\caption{Phase 3 regression predicting iteration rounds and satisfaction rating between \modelthree\ \& \modelfour. \modelfour\ achieved satisfying results in significantly fewer rounds (1.35 fewer on average) than \modelthree\, while generating significantly more satisfying results (higher by 0.65 on a 7-point Likert scale).}
\vspace{-1mm}
\label{tab:phase3}
\end{table}
\subsubsection{Prompting Patterns - Phase 2 and 3}
Our regression results in Table~\ref{tab:phase2} reveal that \modeltwo\ was able to help users achieve satisfiable results in fewer rounds than \modelone\ in Phase 2 and generated significantly more satisfiable images overall. Similarly, in Phase 3 (see Table~\ref{tab:phase3}), \modelfour\ achieved satisfying results in significantly fewer rounds (1.35 on average) than \modelthree\, while generating significantly more satisfying results.
Next, we summarize qualitative insights for participants' prompting patterns throughout Phases 2 and 3. 
Note that a cross-panel analysis of Phase 2 and 3 is in the Appendix, while Figure \ref{fig:first_round} illustrates the association between perceived image diversity and speed in reaching satisfaction.

\begin{figure}[t]
\centering
\subcaptionbox{Phase2}{\includegraphics[width=0.50\linewidth]{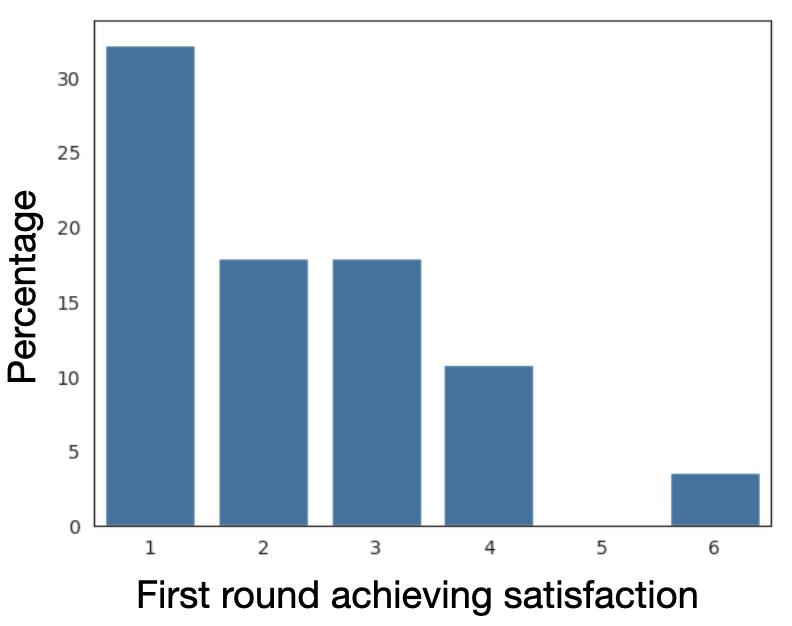}}%
\hfill
\subcaptionbox{Phase3}{\includegraphics[width=0.50\linewidth]{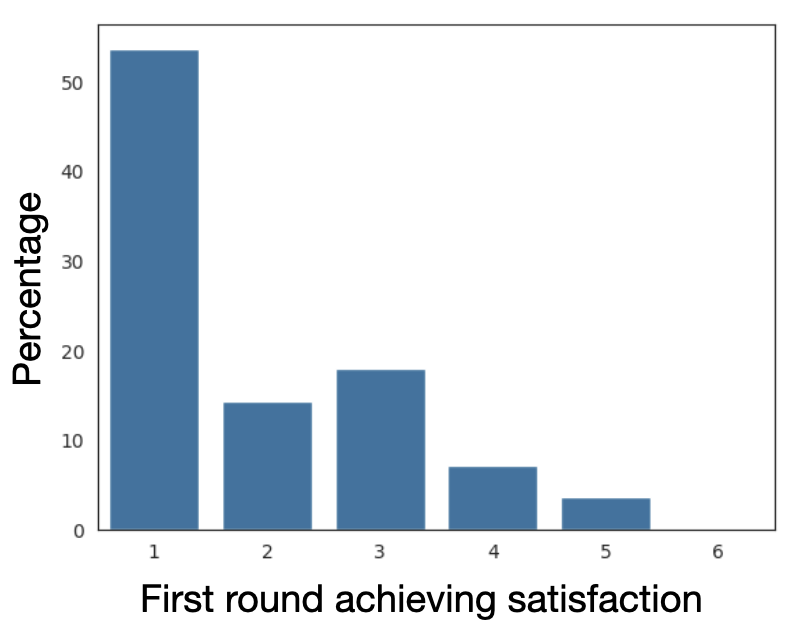}}%
\caption{Cross-Phase Analysis: Participants who perceived the experimental condition with the \model~model as generating more diverse images than the baseline were more likely to reach satisfaction with the first set of images generated by \model~ in subsequent phases.}
\label{fig:first_round}
\end{figure}

\paragraph{Pushing Normative Boundaries to Approach Unique \& Novel Ideas}

\begin{figure}[t]
\includegraphics[width=0.50\textwidth]{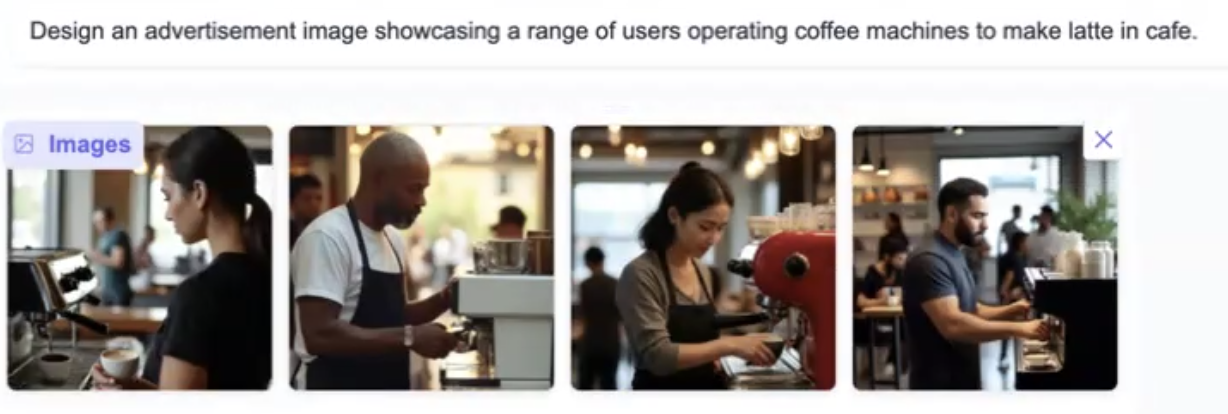}

\caption{Prompt and result images where P26 appreciated age variation but developed a new criteria for use cases/places during the experimental condition of Phase 3.}
\label{fig:p26}
\end{figure}

We saw participants prompt in ways that showed them going beyond normative boundaries in order to generate images that felt novel or surprising. In her first prompt, P28 indicated she wanted to \textit{``have an image with both genders.''} In fact, many participants expressed their satisfaction when the generated images demonstrated visible conceptual differences during the experimental condition (e.g., P2, P17) and dissatisfaction when the images did not offer the variety along the dimensions they expressed desires for during the baseline condition (e.g., P10, P28). 
When the model generated images that varied in the ways 
that matched the participants' expectations or desires, participants immediately developed more criteria for new ways of showcasing diversity among images. 
For instance, P26 appreciated how images generated from his revised prompt showed diversity in age --- \textit{``aged customers, like old [people], teenagers, young man, a young girl''} (see Figure \ref{fig:p26}) --- but expressed immediately after that he wanted to also see these users \textit{``using the coffee machine [in] different places and [use] cases.''} 

\begin{figure}[t]
\includegraphics[width=0.50\textwidth]{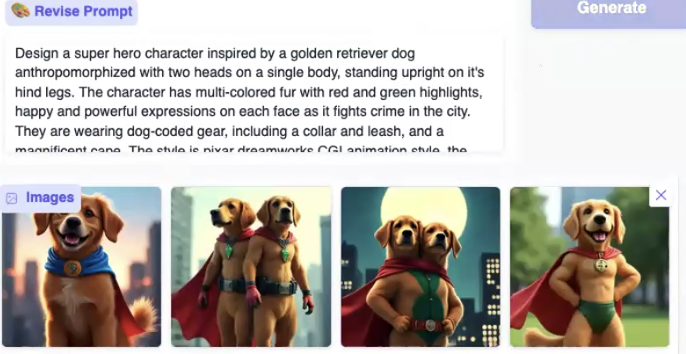}

\caption{ P8 decided to ``go crazy'' by entirely rewriting the prompt with imaginative superhero attributes for S3 in the experimental condition of Phase 3.
}
\label{fig:p8}
\end{figure}

In later prompt iterations, participants explored a variety of techniques. Some (e.g., P8, P10) started generating new prompts entirely as opposed to iteratively modifying the initial prompt --- P8 chose to \textit{``get a little crazy with this''} (see Figure \ref{fig:p8}); others refined their prompts by using more polite or suggestive language to emphasize specific attributes they wanted to see (e.g., P1, P18, P24), and yet others experiment with novel formatting of prompts --- P17 ended one task with a prompt format that included a bulleted list of expected criteria, while P1 tried wrapping a phrase in the * character to emphasize an unfollowed criteria. 

\begin{figure}[t]
\includegraphics[width=0.50\textwidth]{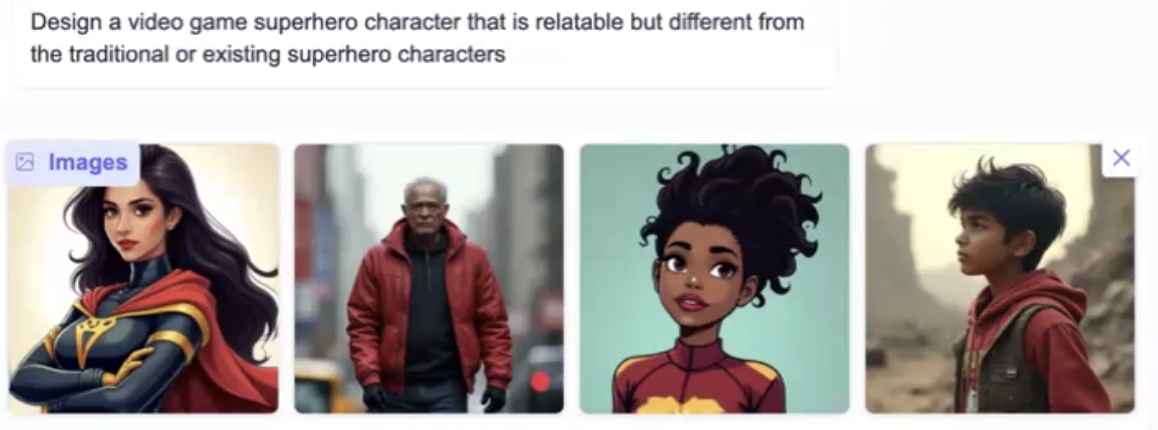}

\caption{ Several participants attempted to break out of the ``Superman'' norm for fictional image generation (S3). Including P4, who found her most satisfied (rightmost) image through the above iteration of prompting during the experimental condition of Phase 3.
}
\label{fig:p4}
\end{figure}

\revised{During the personalization condition, participants (P13, 27) reflected more on their own subjective ``personal bias'' and ``preference[s]'', and instead selected} images with ``unique'' aspects that either differed from their normative expectations or deviated from images generated in previous rounds, and many participants expressed satisfaction with unique aspects of images generated in the experimental conditions (e.g., P7, P12, P18, P24, P27). Most compelling is the example of participants interacting with the fictional character generation scenario (S3): some participants shared how \textit{``you want your video game character to look unique and not like something that already exists''} (P12), others like P4 made directly requested characters who were \textit{``different from the traditional''} (see Figure \ref{fig:p4}). In the case of Phase 3, participants like P18 and P24 explicitly selected images as their favorite because of their unique attributes. P26 eventually reached a high level of satisfaction after multiple rounds of prompting, describing that \textit{``the design is out of [his] imagination. [He] didn't expect this output [and] think[s] `Oh, that's really amazing.' ''} 

However, not all participants sought to break away from norms. A few leaned further into conventional expectations, motivated by the belief that familiar or broadly recognizable imagery would appeal to a wider audience. For instance, P13 explained, \textit{``for a wide range of users, [she]’d probably pick one of the more male-presenting superheroes''} reasoning that such characters were more likely to resonate due to their cultural familiarity. \revised{With the personalized condition, participants also found satisfaction in results catering to their preferences, such as an inclination toward ``friendlier and \dots less aggressive'' images (P27).}


\paragraph{From Exploration to Specificity}

In Phases 2 and 3, participants' prompting patterns revealed an overall shift in intent --- starting from an exploratory attitude and eventually shifting to more specific desired criteria.
Many began the prompting exercise with the intention of openly exploring potential options. These participants approached initial iterations of prompting with a sense of curiosity, without a clear set of developed criteria of what they wanted in an image. For example, P18 described that she was \textit{``curious to see what [the model will] generate\dots so [she] would keep [her criteria] open at first.''} Through later rounds of prompting, several participants developed a taste for more specific criteria to be met in generated images. This pattern applied across all four scenarios in examples such as criteria for a \textit{``rainy''} and \textit{``cozy''} style (P27) for the product advertisement scenario and a \textit{``specific breed of dog''} and \textit{``plants''} in the kitchen and living room for the interior design scenario (P21). 

% \todo{I feel the following one is hard to follow, if not critical, we can consider removing? Jane: wait maybe we just need to reframe}  
However, as participants developed more specific criteria in their prompting, they also started hitting new limitations with model capabilities, requiring them to find creative workarounds to fine-tune prompts in a way that more directly expresses and defines latent (and sometimes explicit) specifications.
% they sometimes also became less satisfied with the images generated -- potentially due to limitations in either their own prompting abilities or the capabilities of the image generation model. 
For example, several participants (e.g., P8, P27) struggled with generating images that omitted certain aspects, using ``negative request[s]'' in their prompts only to see the factors they wanted to exclude, such as getting overwhelmingly Superman-like features when the prompt includes the phrase \textit{``Do not just use superman as an example character''} (P12). In another case, P24 gave the model \textit{``two different options [of]\dots doing different activities like jet skiing or on a cruise ship''}, and while the image generated had boats, it did not include a cruise ship or a jet ski.

\paragraph{Tradeoffs with Realism \& Prompt Fidelity}
For some participants, the tradeoff between how realistic an image appeared versus its adherence to their specified criteria was a challenge to navigate. For some, realistic images were more desirable because they wanted to use real images to inspire their prompting (P28) or viewed realistic images as more appropriate for the situation (P24). Others valued realism as a way to counter exaggerated stereotypes, for example, P3 noted that illustrative images often overemphasized the breasts of female-presenting characters. Several participants expressed dissatisfaction with images that were not photo-realistic (e.g., P3, P24). P24, for example, associated a lack of photorealism with poor image design, stating that if \textit{``you're trying to attract people to go to this place, and this feels AI-generated \dots the image looks like you just cropped it and threw it in there like it wasn't well-made.''} 

On the other hand, even if images were photorealistic, some participants (e.g., P6, P14, P26) were unsatisfied when they depicted unrealistic situations --- such as an initial image set for S4 where users gathered around for a picnic outside at a park.

% \begin{figure*}[]
% \centering
% \subcaptionbox{Phase2 - Preference Rating}{\includegraphics[width=0.50\textwidth]{figures/charts/preference-boxplot.png}}%
% \hfill
% \subcaptionbox{Phase3 - Round Number of Favorite Image}{\includegraphics[width=0.50\textwidth]{figures/charts/favorite-round.png}}%
% \caption{User Study: Phase 2\&3-Image Preference. \todo{explain a bit more and make caption more exciting and summarize key message}  }
% \end{figure*}

% Evaluating personalization: 
% \begin{itemize}
%     \item Re-prompt round number
%     \item Satisfactory rating
%     \item Phase2 - final preferred image rating 
%     \item Phase3 - liked/disliked image round number
% \end{itemize}

% Visualization:
% \begin{itemize}
%     \item textual inversion examples, showing different common spaces found (e.g., background, actions, etc)
%     \item filtered images examples
%     \item experimental vs. baseline generated images (diverseness -- as part of teaser image) 
%     \item comparing how personalize change user prompts/images generation (phase2 vs phase3)
%     \item user-favorite image change across rounds
% \end{itemize}

\section{Discussion}
% \begin{itemize}
    % \item certain use cases need more diverse generation for exploration than others, for instance product advertisement or maybe even interior design
    % \item part of our method naturally leads users to have negative requests (when they see expressions along dimensions that are in conflict with their desires or needs). the inability to take negative prompts remains an open issue with our model and generative t2i more broadly -- perhaps room to think about ways to go around it in future designs? 
    % \item while output spaces can be further expanded, implicit social biases are more difficult to overcome - potentially limiting the extent of expansions (right like some people still chose male characters because they believed that would be "widely appealing")
%     \item user prompts also influence homogenization, something we can't control \cite{patterns} -- opportunity for us to list previous literature on prompt expansion
% \end{itemize}

% \paragraph{Context Shaping Perception of Diversity.}
% Our findings indicate that certain creative contexts may inherently demand greater diversity in generative outputs to effectively support ideation. 

Our findings surface several insights and open issues that inform the future design and development of generative text-to-image systems. Drawing from both quantitative and qualitative analysis, we highlight key considerations that designers and researchers should consider when examining how we can improve systems to be more effective in supporting users' creative goals.

\paragraph{Limit of Technical Expansion.}
Although \model~ successfully expanded generative output spaces along diverse visual dimensions, we observed several explicit limitations inherent in current generative text-to-image models. Notably, participants faced challenges when attempting negative prompting, and they frequently attempted to (usually unsuccessfully) exclude certain features in images or ``negative prompting'' --- highlighting an unresolved limitation in current text-to-image systems \cite{negative} and causing significant frustration. For instance, when they explicitly asked for more variety to overcome normative options such as in the character generation scenario (S3), the model still produced only superman-like designs. 
% The challenge our participants faced with negative prompting highlights an unresolved limitation in current text-to-image systems \cite{negative}, including our own. 
While our findings revealed how exposure to undesired or stereotypical outputs prompted users to explicitly try negating certain attributes, the failures of models to adhere to these requests reveal an important gap between user goals and system performance. Current interaction paradigms may be predominantly optimized for affirmative prompts \cite{affirmative}, inadequately addressing the nuances preferences of users aiming to filter or exclude undesirable elements \cite{positivity}. Addressing this issue presents an interesting opportunity for future text-to-image system design of methods capable of understanding and effectively operationalizing user intentions expressed with negatives/negations.

Additionally, expanding diversity often leads to creative trade-offs with realism or prompt fidelity --- outputs sometimes become less realistic, inconsistent, or semantically distant from user prompts, negatively impacting the perceived quality. Notably, decreased realism in generative outputs sometimes led participants to perceive images as lower quality or less thoughtfully produced, suggesting a critical consideration for future expansions of generative diversity to preserve perceived quality alongside increased variety \cite{diverse_quality, vqvae}. Our analysis revealed that several participants were bothered by the style of images feeling inauthentic because they did not seem photo-realistic -- an established challenge in AI-generated images \cite{aigc}. Participants also encountered difficulties in exerting fine-grained control over specific visual attributes, such as precisely defined character traits or detailed stylistic elements. Moreover, implicit biases embedded in the training data of generative models themselves limited genuine diversity, often resulting in the continued generation of stereotypical or normative outputs even when expansions were explicitly fine-tuned and prompted for. Overall, we note these technical boundaries reveal important avenues for future improvements in the state-of-the-art for text-to-image generation systems and emphasize the need for methods that better accommodate these challenges.

\paragraph{How User Expectations and Behaviors Shape Diversity.}
Our findings reveal an interesting interplay between user expectations and prompting behaviors, which jointly shape how diversity was valued and pursued in different creative scenarios in our user study. While generative systems inherently offer the potential to expand creative exploration along multiple dimensions, participants' preferences toward diversity varied considerably based on implicit normative ideas, which in turn influences the visual diversity of AI-generated images \cite{patterns}. For example, in the interior design scenario (S4), several participants explicitly preferred outputs that adhered to culturally dominant aesthetics, such as Western minimalism, thus implicitly limiting the extent to which they engaged with more diverse dimensions of images. In contrast, in the tourist promotion (S2) and fictional character generation (S3) scenarios, many participants actively resisted normative stereotypes, explicitly prompting for more unique and non-normative outputs such as those different from known superheroes with diverse demographic identities. 

Participants' selective attention to prominent visual dimensions, such as gender or environment, significantly influenced their perception of diversity. This selective focus often led participants to overlook subtler yet equally important elements of diversity in the generated images. Our analysis also highlights how user-generative prompts themselves often reinforced certain normative boundaries--potentially contributing to output homogenization independent of a model's technical constraints \cite{chinchure2024tibet, openbias, lens, eval, stablebias}. Some participants inadvertently constrained the diversity of images generated by embedding implicit stereotypes in their prompts, even when presented with more expansive generative capabilities, for purposes of catering to the more common audiences. 
Interestingly, a few of these participants were actively aware of such assumptions they were making but still felt that images relevant to the scenario had to meet certain normative criteria -- e.g.,  that a male-presenting superhero character will be better perceived because it reflects the societal representation of superheroes. This practice also illustrates how prompting might serve a mediating role between users and models: while the generative system can support diverse output spaces, the users' own prompting approach can still significantly limit or promote exploratory ideation. Notably, this could reveal a critical limitation in that user input remains a bottleneck that models cannot easily control.

\section{Limitations \& Future Work}
Our user study highlighted common limitations in current image generation systems. Participants noted mismatches between prompts and images—like the wrong number of people or hands—as well as visual artifacts such as distorted fingers or unrealistic scenes. These are known issues in pre-trained text-to-image models and remain open challenges~\citep{liang2024richhumanfeedbacktexttoimage, han2025progressivecompositionalitytexttoimagegenerative}. We believe future work should continue to improve compositional alignment and visual quality. 
% Due to limited resources, o
Our study was not conducted at a large scale -- future work expanding to more diverse demographics (e.g., race, education, and culture) could offer broader insights and more generalizable results. Future directions include:
\paragraph{Fine-tuning for personalization with preference data to further mitigate observed tradeoffs with fidelity.} Our system enables the study of user preference data supporting creative and personalized image generation, which can further fine-tune pre-trained text-to-image models. The form of feedback we study is richer than typical post-training paradigms in generative models since it includes both preferences about currently seen prompts and images as well as their expanded versions. We expect this personalization to help further balance visual diversity and semantic fidelity, allowing the system to adapt to fine-grained and individual user preferences over time.
\paragraph{Extension to generating video, audio, and other forms of creative media.} Generating diverse outputs is harder in video or audio than in image generation, as models must maintain both spatial and temporal coherence. Expanding the framework to handle temporally or acoustically coherent modalities, such as video and audio, would unlock new applications in storytelling, music composition, and multimodal content creation. Our proposed approaches in automatic dimension identification, expansion, and personalization can lead to more creative media generation.
\paragraph{New user interfaces that communicate model decision making (explainability) and enable fine-grained intervention/control} Another promising direction is to design user interfaces that support richer interaction with model outputs, allowing users to understand how decisions are made and to provide fine-grained feedback or control. Future work can augment text-to-image models to perform step-by-step generation of image regions while explaining their internal reasoning process. By developing new user interfaces that convey the model's internal language explanations and take into account real-time user specifications, we can adaptively improve their generation quality and diversity for different preferences.

\section{Conclusions}
In this work, we contribute \model, a text-to-image generation tool that supports creative professionals in the early ideation stages of their work by automatically expanding their visual output spaces and leveraging user feedback to personalize to preferences in real-time. Through a user study that engaged 28 participants in creative tasks, \model~ not only enabled users to achieve more satisfiable results in fewer prompt iterations, it also helped them notice and push back against normative values embedded in current models, as well as construct and define novel specifications around desired outputs.

\begin{acks}
This work was partially supported by NSF DGE2140739. We would also like to thank the Human–Computer Interaction Institute at Carnegie Mellon University and the MIT Media Lab for their valuable guidance, facilities, and feedback during the course of this research.
\end{acks}

\bibliographystyle{ACM-Reference-Format}
\bibliography{references}

\end{document}